%% file: main_rss_2020.tex
\documentclass[conference]{IEEEtran}
\usepackage{times}

\usepackage{graphicx}
\usepackage{color}
\usepackage{subcaption}
\usepackage{amsmath}
\usepackage{amsfonts}
\usepackage{algorithm}
\usepackage[noend]{algpseudocode}
\usepackage{tabularx}
\usepackage[section]{placeins}

\newcommand{\E}{\mathbb{E}}
\newcommand{\T}{\mathcal{T}}
\newcommand{\act}{\mathbf{a}}
\newcommand{\s}{\mathbf{s}}
\newcommand{\e}{e}

\newcommand\blfootnote[1]{%
  \begingroup
  \renewcommand\thefootnote{}\footnote{#1}%
  \addtocounter{footnote}{-1}%
  \endgroup
}

\usepackage[numbers]{natbib}
\usepackage{multicol}
\usepackage[bookmarks=true]{hyperref}

\usepackage[symbol]{footmisc}

\begin{document}

% paper title
\title{MT-Opt: Continuous Multi-Task Robotic\\ Reinforcement Learning at Scale}

\author{
\authorblockN{Dmitry Kalashnikov*, Jacob Varley*, Yevgen Chebotar, Benjamin Swanson,\\ Rico Jonschkowski, Chelsea Finn, Sergey Levine, Karol Hausman*}\\ 
\authorblockA{Robotics at Google}
\vspace{-10pt}
}

\maketitle

\input{sections/abstract}

\IEEEpeerreviewmaketitle

\input{sections/introduction}

\input{sections/related_work}

\input{sections/system_overview}

\input{sections/method}

\input{sections/data_collection}

\input{sections/real_experiments}

\input{sections/conclusion}
\input{sections/acknowledgements}

\bibliographystyle{plainnat}
\bibliography{references}

\input{sections/appendix}

\end{document}

%% file: sections/abstract.tex
\begin{abstract}
General-purpose robotic systems must master a large repertoire of diverse skills to be useful in a range of daily tasks. 
While reinforcement learning provides a powerful framework for acquiring individual behaviors, the time needed to acquire each skill makes the prospect of a generalist robot trained with RL daunting. In this paper, we study how a large-scale collective robotic learning system can acquire a repertoire of behaviors simultaneously, sharing exploration, experience, and representations across tasks. 
In this framework new tasks can be continuously instantiated from previously learned tasks improving overall performance and capabilities of the system.
To instantiate this system, we develop a scalable and intuitive framework for specifying new tasks through user-provided examples of desired outcomes, devise a multi-robot collective learning system for data collection that simultaneously collects experience for multiple tasks, and develop a scalable and generalizable multi-task deep reinforcement learning method, which we call MT-Opt. 
We demonstrate how MT-Opt can learn a wide range of skills, including semantic picking (i.e., picking an object from a particular category), placing into various fixtures (e.g., placing a food item onto a plate), covering, aligning, and rearranging. 
We train and evaluate our system on a set of 12 real-world tasks with data collected from 7 robots, and demonstrate the performance of our system both in terms of its ability to generalize to structurally similar new tasks, and acquire distinct new tasks more quickly by leveraging past experience. 
We recommend viewing the videos at \href{karolhausman.github.io/mt-opt}{\color{blue}{\url{karolhausman.github.io/mt-opt}}}
\end{abstract}

%% file: sections/introduction.tex
\section{Introduction}
\label{sec:introduction}
\blfootnote{${^*}$Equal contribution}
Today's deep reinforcement learning (RL) methods, when applied to real-world robotic tasks, provide an effective but expensive way of learning skills~\cite{kalashnikov2018qt,akkaya2019solving}.
While existing methods are effective and able to generalize, they require considerable on-robot training time, as well as extensive engineering effort for setting up each task and ensuring that the robot can attempt the task repeatedly.
For example, the QT-Opt~\cite{kalashnikov2018qt} system can learn vision-based robotic grasping, but it requires over $500,000$ trials collected across multiple robots. 
While such sample complexity may be reasonable if the robot needs to perform a single task, such as grasping objects from a bin, it becomes costly if we consider the prospect of training a general-purpose robot with a large repertoire of behaviors, where each behavior is learned in isolation, starting from scratch. %requires this sort of sample complexity. 
Can we instead \emph{amortize} the cost of learning this repertoire over multiple skills, where the effort needed to learn whole repertoire is reduced, easier skills serve to facilitate the acquisition of more complex ones, and data requirements, though still high overall, become low for each individual behavior?

\begin{figure}[t]
    \centering
    \includegraphics[width=0.99\linewidth]{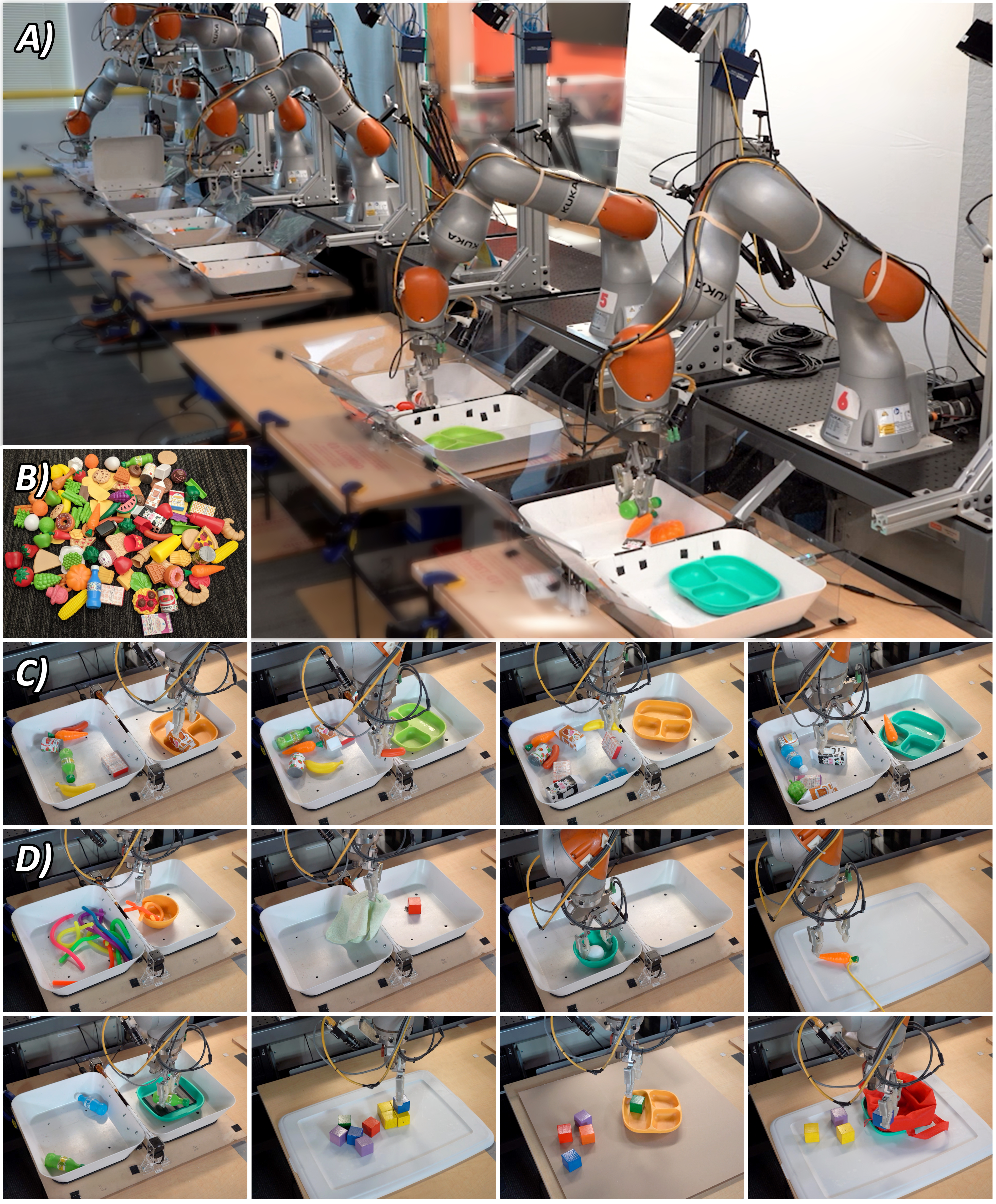}
    \vspace{-12pt}
    \caption{\small A) Multi-task data collection. B) Training objects. C) Sample of tasks that the system is trained on.
    D) Sample of behaviorally and visually distinct tasks such as covering, chasing, alignment, which we show our method can adapt to. MT-Opt learns new tasks faster (potentially zero-shot if there is sufficient overlap with existing tasks), and with less data compared to learning the new task in isolation.}
    \label{fig:system_teaser}
    \vspace{-10pt}
\end{figure}

Prior work indicates that multi-task RL can indeed amortize the cost of single-task learning~\cite{espeholt2018impala,PintoG17,riedmiller2018learning,wilson2007multi,hausman2018learning}. 
In particular, insofar as the tasks share common structure, if that structure can be discovered by the learning algorithm, all of the tasks can in principle be learned much more efficiently than learning each of the tasks individually. %, amortizing the cost of acquiring a skill. 
Such shared representations can include basic visual features, as well as more complex concepts, such as learning how to pick up objects. 
In addition, by collecting experience simultaneously using controllers for a variety of tasks with different difficulty, the easier tasks can serve to ``bootstrap'' the harder tasks. 
For example, the task of placing three food items on a plate may be difficult to complete if the reward is provided only at the end, but picking up a single food item is considerably easier. 
By learning these tasks together, the easier task serves to aid with exploration for the harder task. 
Finally, by enabling the multi-task RL policy to learn shared representations, learning new tasks can become easier over time as the system acquires more skills and learns more widely-useful aspects of the environment.

However, to realize these benefits for a real-world robotic learning system, we need to overcome a number of major challenges~\cite{schaul2019ray,hessel2019popart,chen2017gradnorm,yu2020gradient}, which have so far made it difficult to produce a large-scale demonstration of multi-task image-based RL that effectively accelerates the acquisition of generalizable real-world robotic skills. 
First, multi-task reinforcement learning is known to be exceedingly difficult from the optimization standpoint, and the hypothesized benefits of multi-task learning have proven hard to realize due to these difficulties~\cite{schaul2019ray, yu2020meta}. 
Second, a real-world multi-task learning framework requires the ability to easily and intuitively define rewards for a large number of tasks. 
Third, while all task-specific data could be shared between all the tasks, it has been shown that reusing data from non-correlated tasks can be harmful to the learning process~\cite{eysenbach2020rewriting}.
Lastly, in order to receive the benefits from shared, multi-task representation, we need to significantly scale up our algorithms, the number of tasks in the environment, and the robotic systems themselves.

The main contribution of this paper is a general multi-task learning system, which we call MT-Opt, that realizes the hypothesized benefits of multi-task RL in the real world while addressing some of the associated challenges. We further make the following contributions:
\begin{itemize}
    \item We address the challenge of providing rewards by creating a scalable and intuitive success-classifier-based approach that allows to quickly define new tasks and their rewards.
    \item We show how our system can quickly acquire new tasks by taking advantage of prior tasks via shared representations, novel data-routing strategies, and learned policies.
    \item We find that, by learning multiple related tasks simultaneously, not only can we increase the data-efficiency of learning each of them, but also solve more complex tasks than in a single-task setup. 
\end{itemize}
We present our multi-task system as well as examples of some of the tasks that it is capable of performing in Fig.~\ref{fig:system_teaser}.

%% file: sections/related_work.tex
\section{Related Work}
\label{sec:related_works}

Multi-task learning, inspired by the ability of humans to transfer knowledge between different tasks~\cite{caruana_multitask_1997}, is a promising approach for sharing structure and data between tasks to improve overall efficiency.
Multi-task architectures have been successful across multiple domains, including applications in natural language processing~\cite{subramanian2018learning, johnson2017google, mccann2018natural,LiuHCG19} and computer vision~\cite{chen2018gradnorm,misra2016cross,sener2018multi,perez2017film,taskonomy2018,standley2020tasks}.
In this work, we apply multi-task learning concept in a reinforcement learning setting to real robotic tasks -- a combination that poses a range of challenges.

Combining multiple task policies has been explored in reinforcement learning by using gating networks~\cite{TaylorS07, MullingKKP13}, conditioning policies on tasks~\cite{DeisenrothEPF14}, mapping tasks to parameters of a policy~\cite{SilvaKB12,KoberWOP12,Yang2020},  distilling separate task policies into a shared multi-task policy~\cite{levine2016end,teh2017distral, rusu2015policy, parisotto2015actor, ghosh2017divide,arora2018multi}.  In this work, we concentrate on directly learning a shared policy to take advantage of the shared structure which as we find in our experiments significantly improves the training efficiency. Advantages of multi-task learning for visual representations has been explored in~\cite{PintoGHPG16}. Similar to our method, Pinto and Gupta~\cite{PintoG17} use a shared neural network architecture for multi-task learning with shared visual layers and separate task-specific layers that are trained with task-specific losses. In contrast, in our work, we concentrate on sparse-reward tasks with a common loss structure within a Q-learning framework. Several works explore how to mitigate multi-task interference and conflicting objectives when optimizing a single model for multiple tasks~\cite{hessel2019multi, pcgrad}. In our experiments, we find that better data routing and grouping of tasks training data helps with not only better mitigating conflicting objectives but also improving learning efficiency through data reuse.

Learning complex and composite skills has been addressed through hierarchical reinforcement learning with options~\cite{SUTTON1999181, BartoM03, DanielNP12}, combining multiple sub-tasks~\cite{Barto:04, Dietterich00, MorimotoD01,GhavamzadehM03,shu2017hierarchical,zeng2018learning}, reusing samples between tasks~\cite{LazaricRB08}, relabeling experience in hindsight~\cite{her}, introducing demonstrations~\cite{rahmatizadeh2018vision,xie2019improvisation,hausman2017multi,li2017infogail,singh2020scalable,Shankar2020Discovering,shaoconcept2robot}. A range of works employ various forms of autonomous supervision
to learn diverse skills, e.g. by scaling up data collection~\cite{PintoG16}, sampling suitable tasks~\cite{sharma2017learning} or goals~\cite{NairPDBLL18}   to practice, learning a task embedding space amenable to sampling~\cite{hausman2018learning}, or learning a dynamics model and using model-predictive control to achieve goals~\cite{FinnL16,ebert2018,LinBI19,linearforesight}. 
{\citet{riedmiller2018learning} learn sparse-reward tasks by solving easier auxiliary tasks and reusing that experience for off-line learning of more complex tasks. Their SAC-X framework shares data across tasks to learn and schedule many tasks, which eventually facilitate the complex task.}
In~\citet{cabi2019scaling}, previously collected experience is relabeled with new reward functions in order to solve new tasks using batch RL without re-collecting the data. In our work, we similarly design techniques for reusing experience between related tasks, which helps us to solve long-horizon problems and learn new tasks by training new success detectors without re-collecting the data. We expand on this direction by providing an in-depth analysis of various data-sharing techniques and applying these techniques to a number of complex tasks and large-scale data collection on real robots.

Multi-task learning can also be posed as a form of meta-learning, as we aim to share the knowledge between tasks to accelerate training. Meta-learning has been both combined with imitation learning~\cite{duan2017one, finn2017one, yu2018one,james2018task,bonardi2020learning,ortega2019meta,yu2019one} and reinforcement learning through context space learning~\cite{wang2016learning,duan2016rl, mishra2017simple, rakelly2019efficient,zintgraf2019varibad} and gradient-based optimization~\cite{finn2017model, rothfuss2018promp, houthooft2018evolved, gupta2018meta, mendonca2019guided}.
Finally, continual acquisition of skills can be seen as a form of lifelong or continual learning~\cite{thrun1995lifelong}. Multiple works address lifelong reinforcement learning through specifically designed model structures~\cite{rusu2016progressive, fernando2017pathnet}, constraints on model parameters~\cite{liu2018rotate} and generative memory architectures~\cite{kamra2017deep}. We design our framework such that any amount of offline data can be shared between tasks and new tasks can be continuously added through new success detectors without re-collecting the data, which allows continuous acquisition of new skills.

%% file: sections/system_overview.tex
\section{System Overview}
\label{sec:system_overview}

A high-level diagram of our multi-task learning system is shown in Fig.~\ref{fig:mt_opt_overview}. 
We devise a distributed, off-policy multi-task reinforcement learning algorithm together with visual success detectors in order to learn multiple robotic manipulation tasks simultaneously. 
Visual success detectors are defined from video examples
of desired outcomes and labelling prior episodes (Fig.~\ref{fig:mt_opt_overview}A).  These success detectors determine how episodes will be leveraged to train an RL policy (Fig.~\ref{fig:mt_opt_overview}B). During evaluation and fine-tuning (Fig.~\ref{fig:mt_opt_overview}C), 
at each time step, a policy takes as input a camera image and a one-hot encoding of the task, and sends a motor command to the robot. 
At the end of each episode, the outcome image of this process is graded by a multi-task visual success detector ($SD$) that determines which tasks were accomplished successfully and assigns a sparse reward 0 or 1 for each task. 
At the next step, the system decides whether another task should be attempted or if the environment should be reset.
The above-described setup can scale to multiple robots, where each robot concurrently collects data for a different, randomly-selected task.
The generated episodes are used as offline data for training future policies (Fig.~\ref{fig:mt_opt_overview}D) and are available to improve success detectors.

We develop multiple strategies that allow our RL algorithm to take advantage of the multi-task training setting.
First, we use a single, multi-task deep neural network to learn a policy for all the tasks simultaneously,
which enables parameter sharing between tasks. Second, we devise data management strategies that share and re-balance data across certain tasks.
Third, since all tasks share data and parameters, we use some tasks as exploration policies for others, which aids in exploration.

In order to cope with a large, multi-task dataset, we build on many features of the distributed off-policy RL setup from QT-Opt~\cite{kalashnikov2018qt}, and extend it to leverage the multi-task nature of our data. 
In the following sections, we describe the details of different parts of this large scale, image-based distributed multi-task reinforcement learning based system.

%% file: sections/method.tex
\section{MT-Opt: a Scalable Multi-Task RL System}
\label{sec:mt-opt}

\begin{figure}[t]
    \centering
    \includegraphics[width=0.9\linewidth]{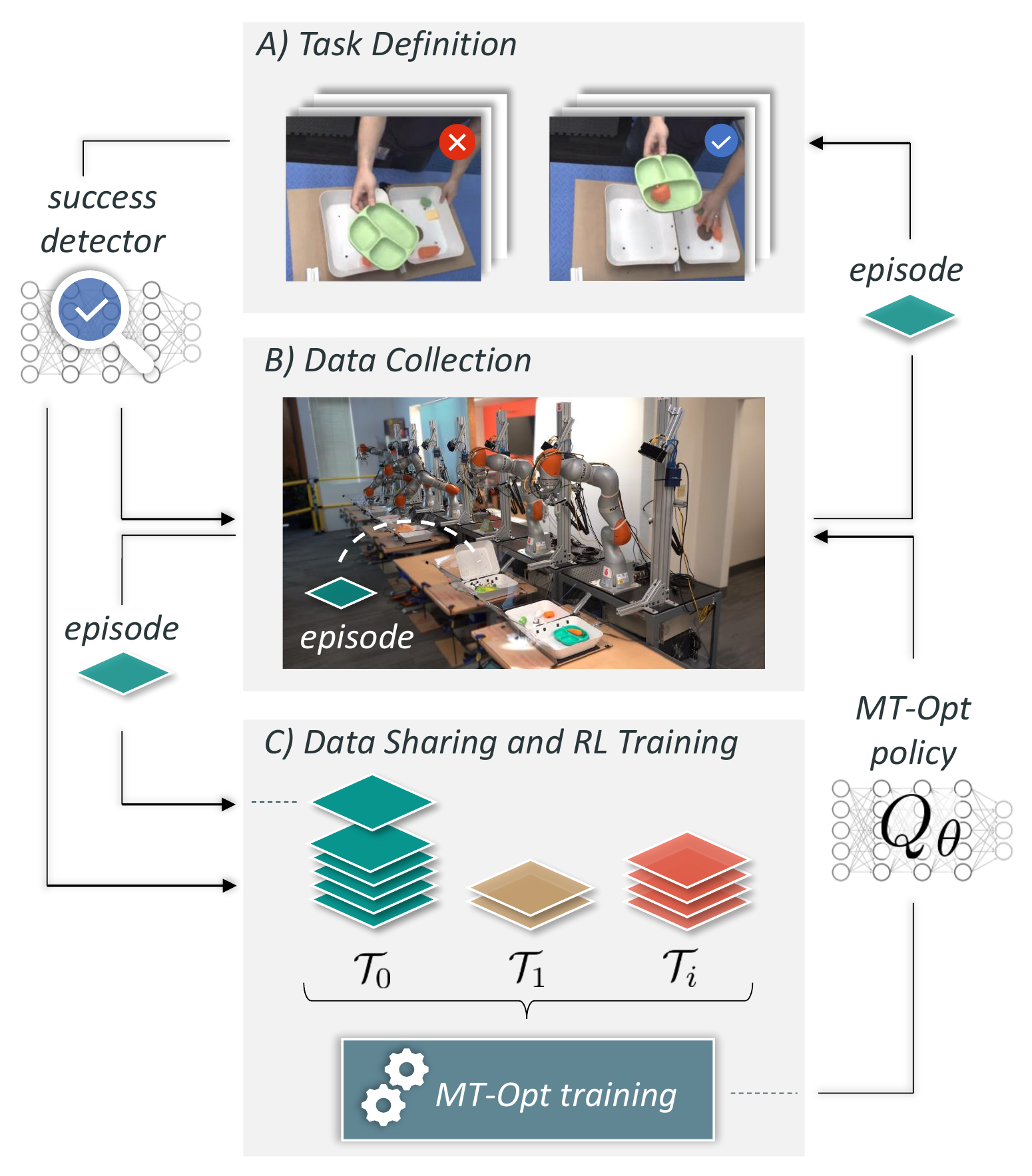}
    \caption{MT-Opt overview. A) The user defines a success detector for tasks through examples of desired outcomes, and relabeling outcomes of prior episodes. B) Utilizing the success detector and the MT-Opt policy, new episodes are collected for multiple tasks. C) Offline episodes enter the data-sharing pipeline that expands and re-balances the data  used to train the MT-Opt policy, while optionally more on-policy data is being collected, particularly for new tasks. This is an iterative process, which results in additional experiences that can be leveraged to define new tasks and train future RL policies.}
    \label{fig:mt_opt_overview}
    \vspace{-10pt}
\end{figure}
In this section, we describe our multi-task reinforcement learning method, MT-Opt, which amortizes the cost of learning multiple skills via parameter and data sharing. 

\subsection{Multi-Task Reinforcement Learning Algorithm}
We first introduce notation and RL fundamentals. We denote the multi-task RL policy as $\pi(\act|\s,\T_i)$,
where $\act \in \mathcal{A}$ denotes the action, which in our case includes the position and the orientation of a robot arm as well as gripper commands, $\s \in \mathcal{S}$ denotes the state, which corresponds to images from the robot's cameras, and $\T_i$ denotes an encoding of the $i^\text{th}$ task drawn from a categorical task distribution $\T_i \sim p(\T)$, which has $n$ possible categories, each corresponding to a different task. At each time step, the policy selects an action $\act$ given the current state $\s$ and the current task $\T_i$ that is set at the beginning of the episode, and receives a task-dependent reward $r_i(\act, \s, \T_i)$. As in a standard Markov decision process (MDP), the environment then transitions to new state $\s'$. The goal of the multi-task RL policy is to maximize the expected sum of rewards for all tasks drawn from the distribution $p(\T)$. The episode finishes when the policy selects a TERMINATE action or reaches a pre-defined maximum step limit.

Our goal is to learn an optimal multi-task Q-Function $Q_\theta(\s, \act, \T_i)$ with parameters $\theta$, that estimates the expected sum of rewards that will be achieved after taking the action $\act$ in the current state $\s$ for the task $\T_i$.
In particular, we build on the single-task QT-Opt algorithm~\cite{kalashnikov2018qt}, which itself is a variant of Q-learning~\cite{sutton98}, and learns a single-task optimal Q-Function by minimizing the Bellman error:
\begin{equation}
    \mathcal{L}_{i}(\theta) = \E_{(\s,\act,\s')\sim p(\s,\act,\s')} {\big[ D (Q_\theta(\s,\act), Q_T(\s, \act, \s')) \big]},
    \label{eq:single_bu}
\end{equation}
where $Q_T(\s,\act,\s') = r(\s, \act) + \gamma V(\s')$
is a target Q-value and $D$ is a divergence metric, such as cross-entropy, $\gamma$ is a discount factor, $V(\s')$ is the target value function of the next state computed using stochastic optimization of the form $V(\s') = \max_{\act'} Q(\s',\act')$, and the expectation is taken w.r.t. previously seen transitions $p(\s,\act,\s')$. Similarly to~\cite{kalashnikov2018qt}, we use the cross-entropy method (CEM) to perform the stochastic optimization to compute the target value function. 

To extend this approach to the multi-task setting, let $\s^{(i)},\act^{(i)}, \s'^{(i)}$ denote a transition that was generated by an episode $\e^{(i)}$ for the $i^\text{th}$  task $\T_i$. As we discuss next, each transition could in fact be used for multiple tasks. In the multi-task case, using Eq.~\ref{eq:single_bu}, the multi-task loss becomes:
\begin{align}
      & \mathcal{L}_\text{multi}(\theta) = \E_{\T_i \sim p(\T)} \left[\mathcal{L}_{i}(\theta) \right] = \E_{\T_i \sim p(\T)} \Big[ \\ 
      &  \E_{p(\s^{(i)},\act^{(i)},\s'^{(i)})}  {\left[ D (Q_\theta(\s^{(i)},\act^{(i)},\T_i), Q_T(\s^{(i)}, \act^{(i)}, \s'^{(i)}, \T_i)) \right]} \Big], \nonumber 
\label{eq:multi_bu}
\end{align}
where $(\s^{(i)},\act^{(i)},\s'^{(i)})$ are transitions generated by tasks $\T_i$.

While this basic multi-task Q-learning system can in principle acquire diverse tasks, with each task learning from the data corresponding to that task, this approach does not take the full advantage of the multi-task aspects of the system, which we introduce next.

\subsection{Task Impersonation and Data Rebalancing}
\label{sec:tasks-impersonation}

\begin{figure}[t]
    \centering
    \includegraphics[width=0.92\linewidth]{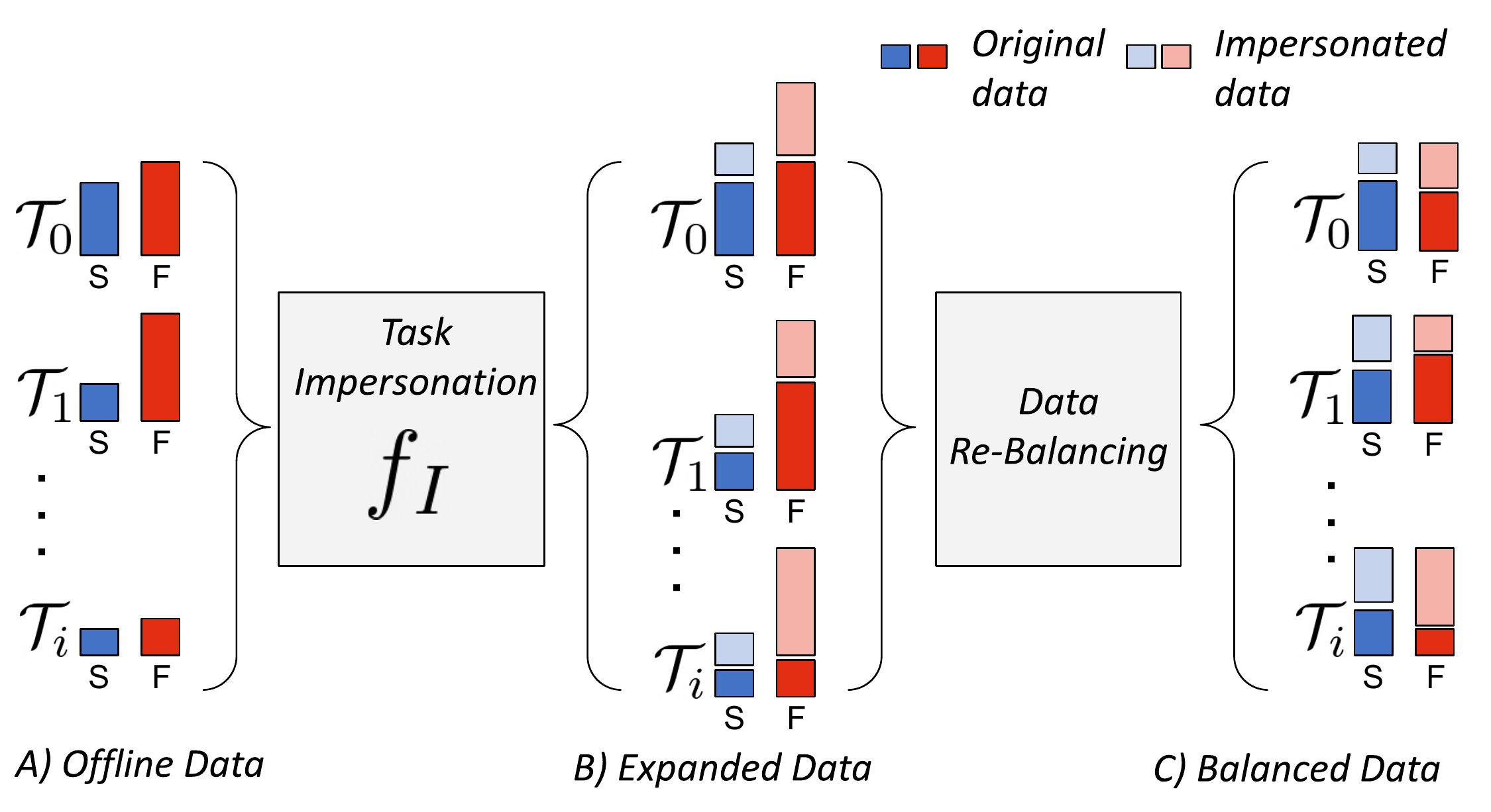}
    \caption{Path of episodes through task impersonation, where episodes are routed to train relevant tasks, and data re-balancing where the ratio of success (S) and failure (F) episodes and proportion of data per task is controlled. Pale blue and pale red indicates additional task training data coming from other tasks. The height of a bar indicates very different amount of data across tasks and across successful outcomes.}
    \label{fig:data_routing}
    \vspace{-7pt}
\end{figure}

One of the advantages of using an off-policy RL algorithm such as Q-learning is that collected experience can be used to update the policy for other tasks, not just the task for which it was originally collected.  This section describes how we effectively train with multi-task data through \textit{task impersonation} and \textit{data re-balancing}, as summarized in Fig.~\ref{fig:data_routing}.

We leverage such experience sharing at the whole episode level rather than at the individual transition level. The goal is to use all transitions of an episode $\e^{(i)}$ generated by task $\T_i$ to aid in training a policy for a \textit{set of $k_i$} tasks $\T_{\{k_i\}}$.
We refer to this process as \textit{task impersonation} (see Algorithm~\ref{alg:impersonation}), where the impersonation function $f_I$ transforms episode data collected for one task into a set of episodes that can be used to also train other tasks, i.e.: $\e^{\{k_i\}} = f_I(\e^{(i)})$. 
Note that in general case $\{k_i\}$ is a \textit{subset} of all tasks $\{n\}$, and it depends on the original task $\T_i$ that the episode $\e^{(i)}$ was collected for.
We introduce this term to emphasize the difference with the \textit{hindsight relabelling}~\cite{andrychowicz2017hindsight} that is commonly used to generate additional successes in a goal-conditioned setting, whereas task-impersonation generates both successes and failures in a task-conditioned setup.

First, we discuss two base choices for the impersonation function $f_I$, then we introduce a more principled solution. 
Consider an identity impersonation function $f_{I_{\text{orig}}}(\e^{(i)}) = \e^{(i)}$, where no task impersonation takes place, i.e. an episode $\e^{(i)}$ generated by task $\T_i$ is used to train the policy exclusively for that task. This baseline impersonation function does not take advantage of the reusable nature of the multi-task data.

At the other end of the data-sharing spectrum is $f_{I_{\text{all}}} = \e^{\{n\}}$, where each task shares data with \textit{all} remaining ${n-1}$ tasks resulting in maximal data sharing. While $f_{I_{\text{orig}}}$ fails to leverage the reusable nature of multi-task data, $f_{I_{\text{all}}}$ can overshare, resulting in many unrelated episodes used as negative examples for the target task. This results in ``dilution'' of intrinsic negatives for a task.
As we will show in Sec.~\ref{sec:mt-opt-quantitative}, this can have disastrous consequences for downstream skill learning.

To address these issues, we devise a new task impersonation strategy $f_{I_\text{skill}}$  that makes use of more fine-grained similarities between tasks.
We refer to it as a skill-based task-impersonation strategy, where we overload the term ``skill'' as a set of tasks that share semantics and dynamics, yet can start from different initial conditions or operate on different objects. For example tasks such as \textit{place-object-on-plate} and \textit{place-object-in-bowl} belong to the same skill.
Our impersonation function $f_{I_\text{skill}}$ allows us to impersonate an episode $\e^{(i)}$ only as the tasks belonging to the same skill as $\T_i$. 
This strategy allows us to keep the benefits of data sharing via impersonation,
while limiting the ``dilution'' issue.
While in this work we manually decide on the task-skill grouping, this can be further extended by learning the impersonation function itself, which we leave as an avenue for future work. In our experiments, we conduct ablation studies comparing $f_{I_\text{skill}}$ (ours) with other task impersonation strategies.

While training, due to the design of our task impersonation mechanism, as well as the variability in difficulty between tasks, the resulting training data stream often becomes highly imbalanced both in terms of the proportion of the dataset belonging to each task, and in terms of the relative frequencies of successful and unsuccessful episodes for each task, see Fig.~\ref{fig:data_routing}B. 
We further highlight the imbalancing challenge in the Appendix, where Fig.~\ref{fig:success_impersonation} shows how much ``extra" data is created per task thanks to the impersonation algorithm. 
In practice, this imbalance can severely hinder learning progress. We found the performance of our system is improved substantially by further re-balancing each batch both between tasks, such that the relative proportion of training data for each task is equal, and within each task, such that the relative proportion of successful and unsuccessful examples is kept constant.

Task impersonation and data re-balancing functions work in sequence and they influence the final composition of the training batch. 
While this process might result in some transitions being drastically oversampled compared to others (if data for that task is scarce), the success and task re-balancing has a big positive impact on the task performance, which we ablate in our experiments.

\begin{figure}[t]
\vspace{-6pt}
\begin{algorithm}[H]
\small
\begin{algorithmic}
	\Procedure{$f_I$}{$e^i$ : original\_episode}
	\State expanded\_episodes = []
	  \State{$SD\{k_i\} \gets$ set of SDs relevant to task $T_i$}
 	  \For {$SD_k$ in $SD\{k_i\}$}
 	    \State\textcolor{blue}{// $e^k$: $e^i$ but with rewards for task $T_k$ not $T_i$}
 	    \State $e^k = SD_k(e^i)$ 
	    \State expanded\_episodes.append($e^k$)
 	  \EndFor 
	\Return expanded\_episodes
	\EndProcedure
	\end{algorithmic}
	\caption{Task Impersonation}
	\label{alg:impersonation}
\end{algorithm}
\vspace{-25pt}
\end{figure}

\section{Rewards via Multi-Task Success Detectors}
\label{sec:sd}

In this work, we aim to learn a discrete set of tasks that can be evaluated based only on the final image of an RL episode. This sparse-reward assumption allows us to train a neural-network-based success detector model ($SD$), which given a final image, infers a probability of a task being successful. 
Similarly to policy learning, we take advantage of the multi-task aspect of this problem and train a single multi-task success detector neural network that is conditioned on the task ID. In fact, we use supervised learning to train a similar neural network architecture model (excluding the inputs responsible for action representation) as for the MT-Opt multi-task policy, which we describe in more detail in the Appendix~\ref{sec:neural_net}. 

To generate training data for the $SD$, we develop an intuitive interface with which a non-technical user can quickly generate positive examples of outcomes that represent success for a particular task. These examples are not demonstrations -- just examples of what successful completion (i.e., the final state) looks like. The user also shows negative examples of near-misses, or outcomes that are visually similar to the positive samples, but are still failures, such as an object being placed next to a plate rather than on top of it. We present example frames of such training data collection process in Fig.~\ref{fig:sd_video_images}.  

While this interface allows us to train the initial version of the multi-task $SD$, additional training data might be required as the robot starts executing that task and runs into states where the $SD$ is not accurate.
Such out of distribution images might be caused by various real-world factors such as different lighting conditions, changing in background surroundings and novel states which the robot discovers. We continue to manually label such images and incrementally retrain $SD$ to obtain the most up-to-date $SD$. In result, we label $\approx 5,000$  images per task and provide more details on the training data statistics in the Appendix, Fig.~\ref{fig:sd_train_data_stats}.

%% file: sections/data_collection.tex
\section{Continuous Data Collection}
\label{sec:data_collection}

\begin{figure}[t]
    \centering
    \includegraphics[width=0.93\linewidth]{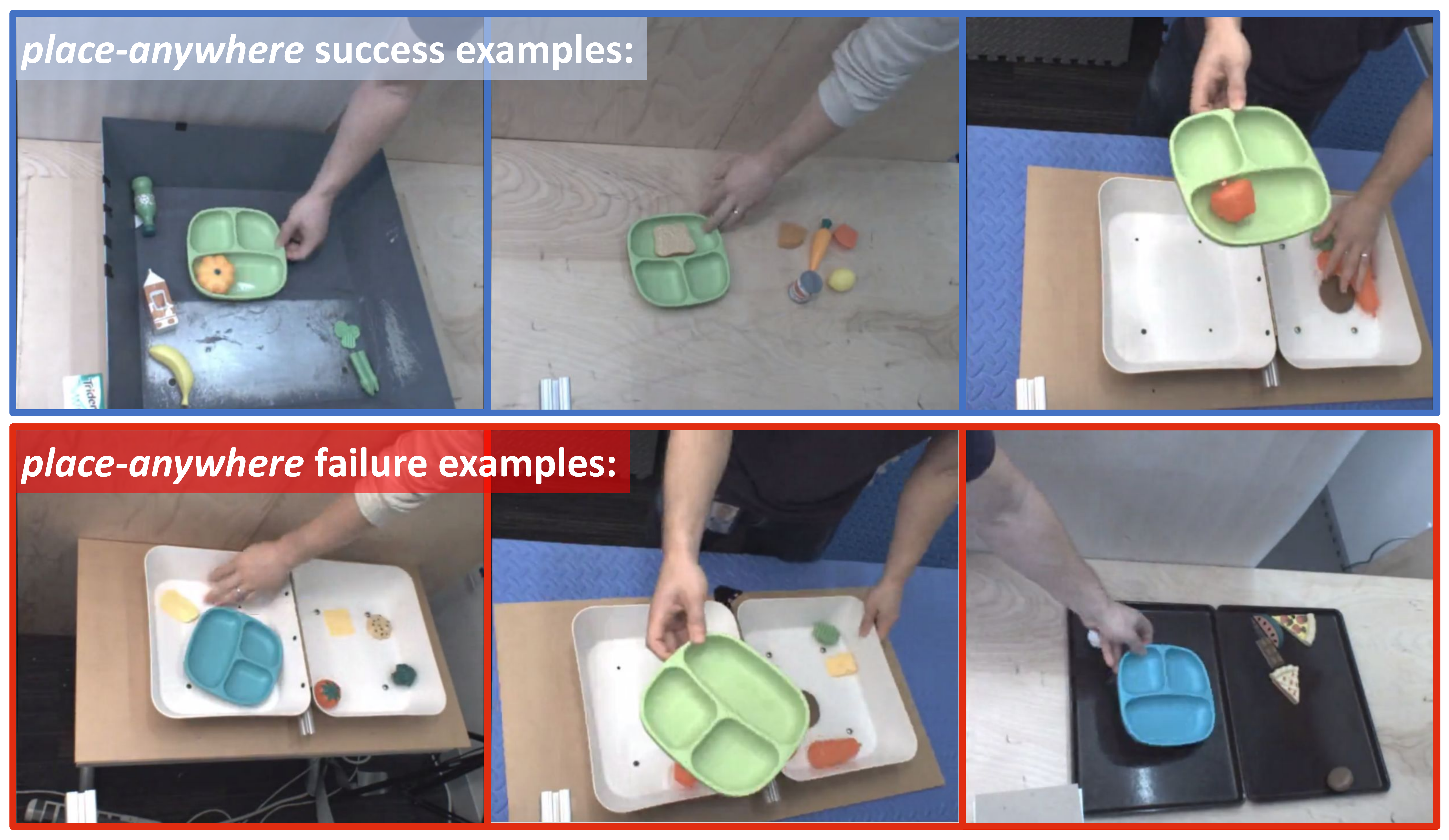}
    \caption{Video frames for the \textit{place-anywhere} task. Success and failure videos are iteratively captured in pairs to mitigate correlations with spurious workspace features such as hands of the user, 
    backgrounds, bins, and distractor objects.}
    \label{fig:sd_video_images}
\end{figure}
\begin{figure}[t]
    \centering
    \includegraphics[width=0.89\linewidth]{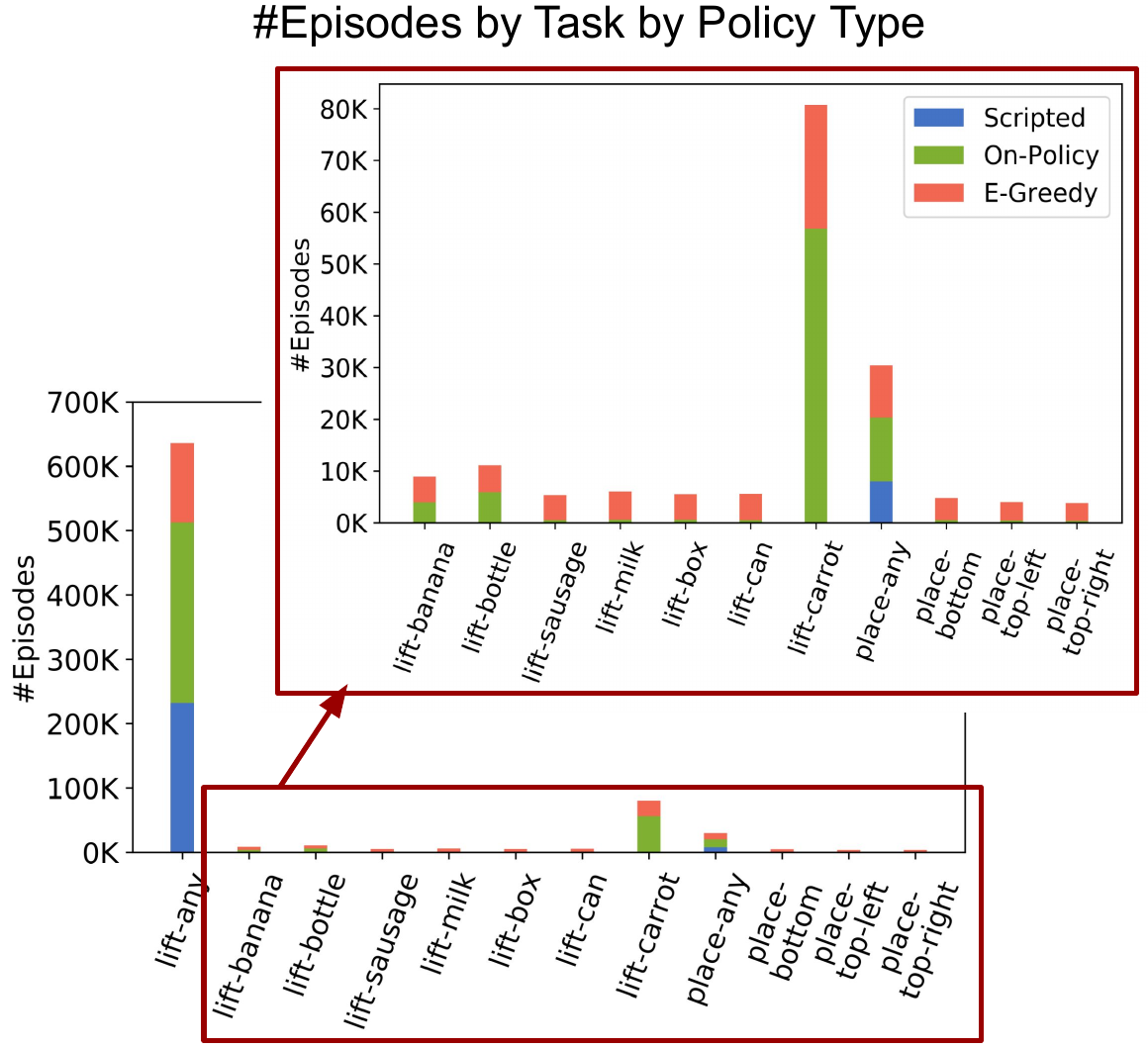}
    \caption{Offline dataset properties. We use our data collection strategy to simultaneously collect data for multiple tasks, where we use easier and more general tasks (e.g. \textit{lift-any}) to bootstrap learning of more complex and specialized tasks (e.g. \textit{lift-carrot}). The resulting multi-task dataset is imbalanced across multiple dimensions:
    the distribution of exploration policies per task (left) and success rate per task (right), both of which we address by using our task-impersonation strategy.}
    \label{fig:data_imbalance}
    \vspace{-5pt}
\end{figure}

\begin{figure*}[t]
    \centering
    \includegraphics[width=0.99\linewidth]{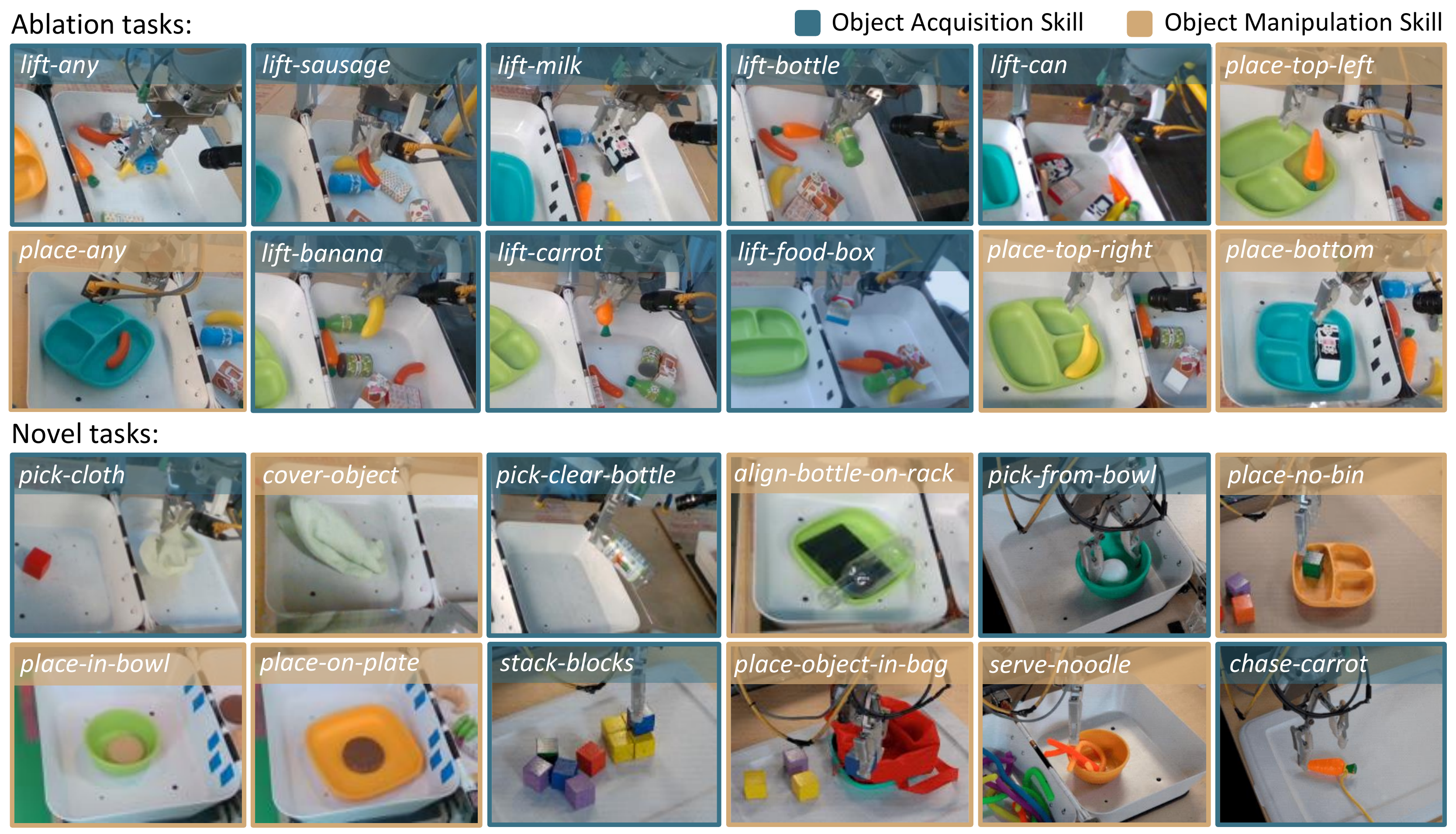}
    \caption{Top: 12 tasks trained for ablations, giving rise to Object Acquisition and Object Manipulation skills. Bottom: examples of additional tasks that a skilled MT-Opt policy can yield occasional success for. These additional tasks can be proactively bootstrapped using MT-Opt as an exploration process and further fine-tuned.
    }
    \label{fig:tasks}
\end{figure*}

In this section, we present the data collection strategy that we utilize to simultaneously collect data for multiple distinct tasks across multiple robots.
Our main observation w.r.t. the multi-task data collection process is that we can use solutions to easier tasks to effectively bootstrap learning of more complex tasks.
This is an important benefit of our multi-task system, where an average MT-Opt policy for simple tasks might occasionally yield episodes successful for harder tasks. Over time, this allows us to start training an MT-Opt policy now for the harder tasks, and consequently, to collect better data for those tasks. 
To kick-start this process and bootstrap our two simplest tasks, we use two crude scripted policies for picking and placing (see Sec.~\ref{subsec:description_of_scripted_policies} for details) following prior work~\cite{kalashnikov2018qt}.
In addition, in order to simplify the exploration problem for longer-horizon tasks, we also allow the individual tasks to be ordered sequentially, where one task is executed after another. 
As such, our multi-task dataset grows over time w.r.t. the amount of per-task data as well as percentage of successful episodes for all the tasks.

Importantly, this fluid data collection process results in an imbalanced dataset, as shown on Fig.~\ref{fig:data_imbalance}. Our data impersonation and re-balancing methods described above address this imbalance by efficiently expanding and normalizing data.

%% file: sections/real_experiments.tex
\section{Experiments}
\label{sec:experiments}

The goal of our real-world experiments is to answer the following questions:
    \textbf{(1)} How does MT-Opt perform, quantitatively and qualitatively, on a large set of vision-based robotic manipulation tasks?  
    \textbf{(2)} Does training a shared model on many tasks improve MT-Opt's performance?
    \textbf{(3)} Does data sharing improve performance of the system? 
    \textbf{(4)} Can our multi-task data collection strategy use easier tasks to bootstrap learning of more difficult tasks?
    \textbf{(5)} Can MT-Opt quickly learn distinct new tasks by adapting learned skills?

\subsection{Experimental Setup}
MT-Opt provides a general robotic skill learning framework that we use to learn multiple tasks, including semantic picking (i.e., picking an object from a particular category), placing into various fixtures (e.g., placing a food item onto a plate), covering, aligning, and rearranging. 
We focus on basic manipulation tasks that require repositioning objects relative to each other. A wide range of manipulation behaviors fall into this category, from simple bin-picking to more complex behaviors, such as covering items with a cloth, placing noodles into a bowl, and inserting bottles into form-fitted slots.
In the following experiments, we use a set of 12 tasks for quantitative evaluation of our algorithm. 
These 12 tasks include a set of plastic food objects and divided plate fixtures and they can be split into `object acquisition' and `object manipulation' skills.
Our most general object acquisition task is \textit{lift-any}, where the goal is to singulate and lift any object to a certain height. In addition, we define 7 semantic lifting tasks, where the goal is to search for and lift a particular object, such as a plastic carrot. 
The placing tasks utilize a divided plate where the simplest task is to place the previously lifted object anywhere on the plate (\textit{place-any}). Harder tasks require placing the object into a particular section of a divided plate, which could be oriented arbitrarily. See Fig.~\ref{fig:tasks} for a visualization of the tasks.

All of the polices used in our studies are trained with offline RL from a large dataset, which we summarize in Fig.~\ref{fig:data_imbalance}.
The resulting policy is deployed on 7 robots attempting each task 100 times for evaluation. 
In order to further reduce the variance of the evaluation, we shuffle the bins after each episode and use a standard evaluation scene (see Appendix, Fig.~\ref{fig:eval_scene}), from which all of the 12 evaluation tasks are feasible.

\subsection{Quantitative and Qualitative Evaluation of MT-Opt}
\label{sec:mt-opt-quantitative}
\begin{figure}[t]
    \centering
    \includegraphics[width=0.97\linewidth]{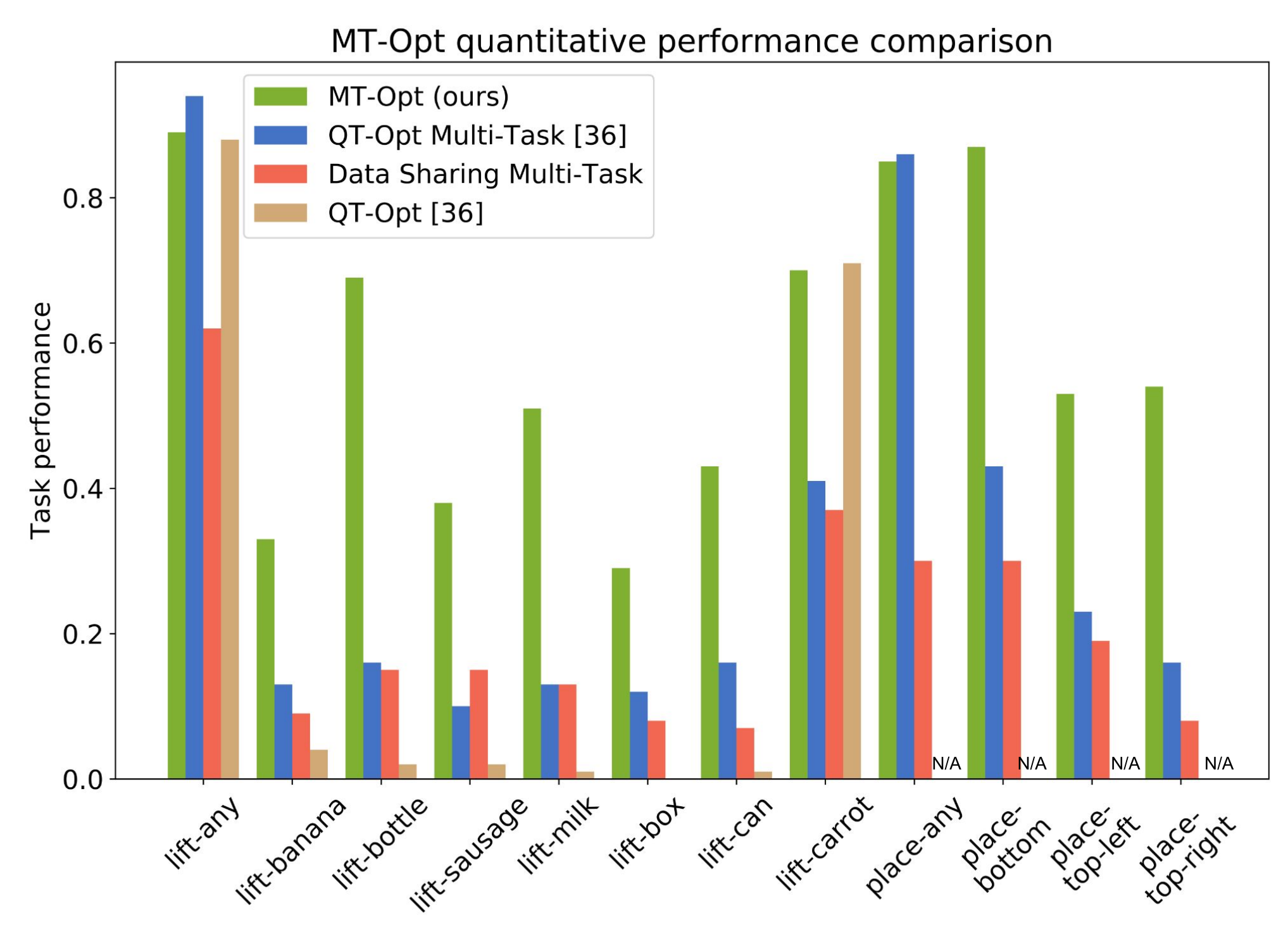}
    \caption{Quantitative evaluation of MT-Opt across 12 tasks. QT-Opt trains each task individually using only data collected for that task. QT-Opt Multi-Task trains a single network for all tasks but does not share the data between them. Data-Sharing Multi-Task also trains a single network for all tasks and shares the data across all tasks without further re-balancing. MT-Opt (ours) provides a significant improvement over the baselines, especially for the harder tasks with less data.}
    \label{fig:quantitative}
\end{figure}

Fig.~\ref{fig:quantitative} shows the success rates of MT-Opt on the 12 evaluation tasks. 
We compare the MT-Opt policy to three baselines: i) single-task QT-Opt~\cite{kalashnikov2018qt}, where each per-task policy is trained separately using only data collected specifically for that task, ii) an enhanced QT-Opt baseline, which we call QT-Opt Multi-Task, where we train a shared policy for all the tasks but there is no data impersonation or re-balancing between the tasks, and iii) a Data-Sharing Multi-Task baseline that is based on the data-sharing strategy presented in~\cite{cabi2019scaling}, where we also train a single Q-Function but the data is shared across all tasks.
Looking at the average performance across all task, we observe that MT-Opt significantly outperforms the baselines, in some cases with $\approx3\times$ average improvement.
While the single task QT-Opt baseline performs similarly to MT-Opt for the task where we have the most data (see the data statistics in Fig.~\ref{fig:data_imbalance}), \textit{lift-any}, its performance drastically drops (to $\approx1\%$) for more difficult, underrepresented tasks, such as \textit{lift-can}. Note, that we are not able run this baseline for the placing tasks, since they require a separate task to lift the object, which is not present in the single-task baseline.
A similar observation applies to QT-Opt Multi-Task, where the performance of rare tasks increases compared to QT-Opt, but is still $\approx4\times$ worse on average than MT-Opt.
Sharing data across all tasks also results in a low performance for semantic lifting and placing tasks and, additionally, it appears to harm the performance of the indiscriminate lifting and placing tasks. 
The MT-Opt policy, besides attaining the $89\%$ success rate on (\textit{lift-any}), also performs the 7 semantic lifting tasks and the 4 placing and rearrangement tasks at a significantly higher success rate than all baselines.
We explain these performance gaps by the way MT-Opt shares the representations and data, and provide a more comprehensive analysis of these factors in the following experiments.
Due to the offline nature of the experiment, this comparison does not take into account the fact that the data for all tasks was collected using the MT-Opt policy. Considering the significantly lower success rates of other methods, it is likely that if the data was collected using these approaches, it would yield much lower success rates, and the gap between MT-Opt and the baselines would further increase.

To further illustrate the learned behavior, we present an example of a successful carrot grasping episode in Fig.~\ref{fig:carrot_and_towel}, where the policy must work the carrot out of the corner before picking it up. 
The challenges of semantic lifting tasks are exacerbated in a small bin setting, where the objects are often crowded and pressed against the walls of the bin, requiring additional actions to retrieve them. 
In addition, we note that semantic picking and placing performance differs substantially between the different tasks. 
While this is due in part to the difficulty of the tasks, this is also caused in large part by the quantity of data available for each object.
To examine this, we refer to Fig.~\ref{fig:data_imbalance}, showing different amounts and type of data collected for various tasks. 
Tasks such as \textit{lift-carrot} and \textit{lift-bottle}, which have more data, especially on-policy data, have higher success rates than underrepresented tasks, such as \textit{lift-box}.
The performance of these underrepresented tasks could be further improved by focusing the data collection on performing them more often.

\subsection{Sharing Representations Between Tasks}
\label{sec:weight-sharing}

To explore the benefits of training a single policy on multiple tasks, we compare the 12-task MT-Opt policy with a 2-task policy that learns \textit{lift-any} and \textit{place-any}.
Both of these policies are evaluated on these two tasks (\textit{lift-any} and \textit{place-any}).
We use the same $f_{I_\text{skill}}$ task impersonation strategy, and the exact same offline dataset (i.e. both policies use the data from the extra 10 narrower tasks, which is impersonated as \textit{lift-any} and \textit{place-any} data) without any on-policy fine-tuning, so data-wise the experiments are identical.

Table~\ref{fig:pick_any_place_any_vs_sota} shows results of the comparison between 12-task and 2-task policies.
The 12-task policy outperforms the 2-task policy \emph{even on the two tasks that the 2-task policy is trained on}, suggesting that training multiple tasks not only enables the 12-task policy to perform more tasks, but also improves its performance on the tasks through sharing of representations.
In particular, the 12-task MT-Opt policy outperforms the 2-task policy by $7\%$ and $22\%$ for the tasks \textit{lift-any} and \textit{place-any}, respectively. 
These results suggest that the additional supervision provided by training on more tasks has a beneficial effect on the shared representations, which may explain the improved performance of the 12-task policy on the indiscriminate lifting and placing tasks.

\begin{table}[t]
\centering
\begin{tabular}{|c|c|c|}
\hline
\multicolumn{3}{|c|}{Parameter Sharing Ablation (Success Rate)} \\ \hline
Model:              & 2-Task Model         & 12-Task Model         \\ \hline
\textit{lift-any}           & 0.82                 & \textbf{0.89}         \\ \hline
\textit{place-any}          & 0.63                 & \textbf{0.85}         \\ \hline
\end{tabular}
    \caption{The effect of parameter sharing: the policy that learns two tasks (\textit{lift-any}, \textit{place any}) in addition to 10 other tasks outperforms a policy trained only for the two target tasks. The two policies are trained from the same offline dataset.}
    \vspace{-7pt}
    \label{fig:pick_any_place_any_vs_sota}
\end{table}

\subsection{Data Sharing Between Tasks}
To test the influence of data-sharing and rebalancing on the multi-task policy's performance, we compare our task impersonation strategy $f_{I_\text{skill}}$ introduced in Sec.~\ref{sec:tasks-impersonation} to a baseline impersonation function that does not share the data between the tasks $f_{I_{\text{orig}}}$, as well as a baseline where each task is impersonated for all other tasks $f_{I_{\text{all}}}$ -- a maximal data sharing strategy.
In our skill-based task impersonation strategy $f_{I_\text{skill}}$, the data is expanded only for the class of tasks having similar visuals, dynamics and goals. 
In addition to $f_{I_\text{skill}}$ task impersonation, we re-balance each training batch between the tasks as well as within each task to keep the relative proportion of successful and unsuccessful trials the same.

The results of this experiment are in Table~\ref{table:data_strategies_table}, with the full results reported in the Appendix, Table~\ref{tab:full_ablation_results}.
Sharing data among tasks using our method of task impersonation and re-balancing provides significant performance improvement across all the evaluation tasks, with improvements of up to $10x$ for some tasks.
The full data-sharing strategy performs worse than both the no-data-sharing baseline and our method, suggesting that \textit{na\"{i}vely sharing all data across all tasks is not effective}.
Because of our data-collection strategy, the resulting multi-task dataset contains much more data for broader tasks (e.g., \textit{lift-any}) than for more narrow, harder tasks (e.g., \textit{lift-box}), as shown in Fig.~\ref{fig:data_imbalance}.
Without any additional data-sharing and re-balancing, this data imbalance causes the baseline strategy $f_{I_{\text{orig}}}$ to attain good performance for the easier, overrepresented tasks, but poor performance on the harder, underrepresented tasks (see Table~\ref{table:data_strategies_table}, first row), whereas our method performs substantially better on these tasks.

\begin{table}[t]
\centering
\begin{tabular}{|c|l|l|}
\hline
\multicolumn{3}{|c|}{Data Strategies Ablation
(min, mean, max, mean of low data tasks)} \\ \hline
Imperson.        & \multicolumn{2}{c|}{Data Re-Balancing Strategy}      \\ \cline{2-3} 
Function             & uniform sampling  & task re-balanced sampling           \\ \hline

$f_{I_{\text{orig}}}$          &      0.10 / 0.32 / 0.94 / 0.18     &           0.16 / 0.55 / 0.85 / 0.42    \\ \hline

$f_{I_{\text{all}}}$                & 0.07 / 0.21 / 0.62 / 0.13  &  0.02 / 0.35 / \textbf{0.95} / 0.21     \\ \hline
$f_{I_{\text{skill}}}$  & 0.17 / 0.46 / 0.88 / 0.32                 & \textbf{0.29} / \textbf{0.58} / 0.89 / \textbf{0.50} (Ours) \\ \hline
\end{tabular}
    \caption{Min, average and max task performance across 12 tasks, as well as average performance across 6 tasks having least data ($\approx6K$ episodes) for different data-sharing strategies. $f_{I_{\text{skill}}}$ impersonation and data re-balancing are complimentary: they both improve over the baselines, while the best effect is achieved by combining both. The effect is especially pronounced for the underrepresented tasks.}
    \label{table:data_strategies_table}
    \vspace{-7pt}
\end{table}
\subsection{Using Easier Tasks to Bootstrap Harder Tasks}
\label{sec:easier-harder}

To explore question (4), we study whether learning an easier but broader task (\textit{lift-any}) can help with a structurally related task that is harder but more specific (\textit{lift-sausage}).
We separate out the data for \textit{lift-sausage} which (as shown in Fig.~\ref{fig:data_imbalance}) consists of $5400$ episodes collected for that task (i.e. $4600$ failures and $800$ successes). In addition, there are $11200$ episodes of successful sausage lifting and as many as $740K$ failures that were collected during the \textit{lift-any} task. 
Combining the \textit{lift-sausage} data and the extra successes from \textit{lift-any} yields $16600$ episodes ($12000$ successes and $4600$ failures). To investigate the influence of MT-Opt and task impersonation on the task-bootstrap problem, we compare our 12-task MT-Opt policy to a single-task policy trained on these $16600$ episodes. These include the exact same set of successful \textit{lift-sausage} episodes as MT-Opt, but does not include the failures from other tasks.

The single-task policy learned from the $16600$ episodes yields performance of $3\%$. 
MT-Opt, which uses impersonated successes and failures, achieves $39\%$ success for the same task, a $\approx 10\times$ improvement.
Both experiments use identical data representing successful episodes.
The benefits of MT-Opt are twofold here. First, we leverage an easier \textit{lift-any} task to collect data for the harder \textit{lift-sausage} task.
Second, the less obvious conclusion can be drawn based on the additional failures impersonated from all other tasks.
This large set of failures, which often include successful grasps of non-target objects, when further re-balanced as described in Sec.~\ref{sec:tasks-impersonation}, results in the significant boost in performance of this task. This demonstrates the value of both successful and unsuccessful data collected by other tasks for learning new tasks.

\subsection{Learning New Tasks with MT-Opt}

\begin{figure}[t]
    \centering
    \includegraphics[width=0.99\linewidth]{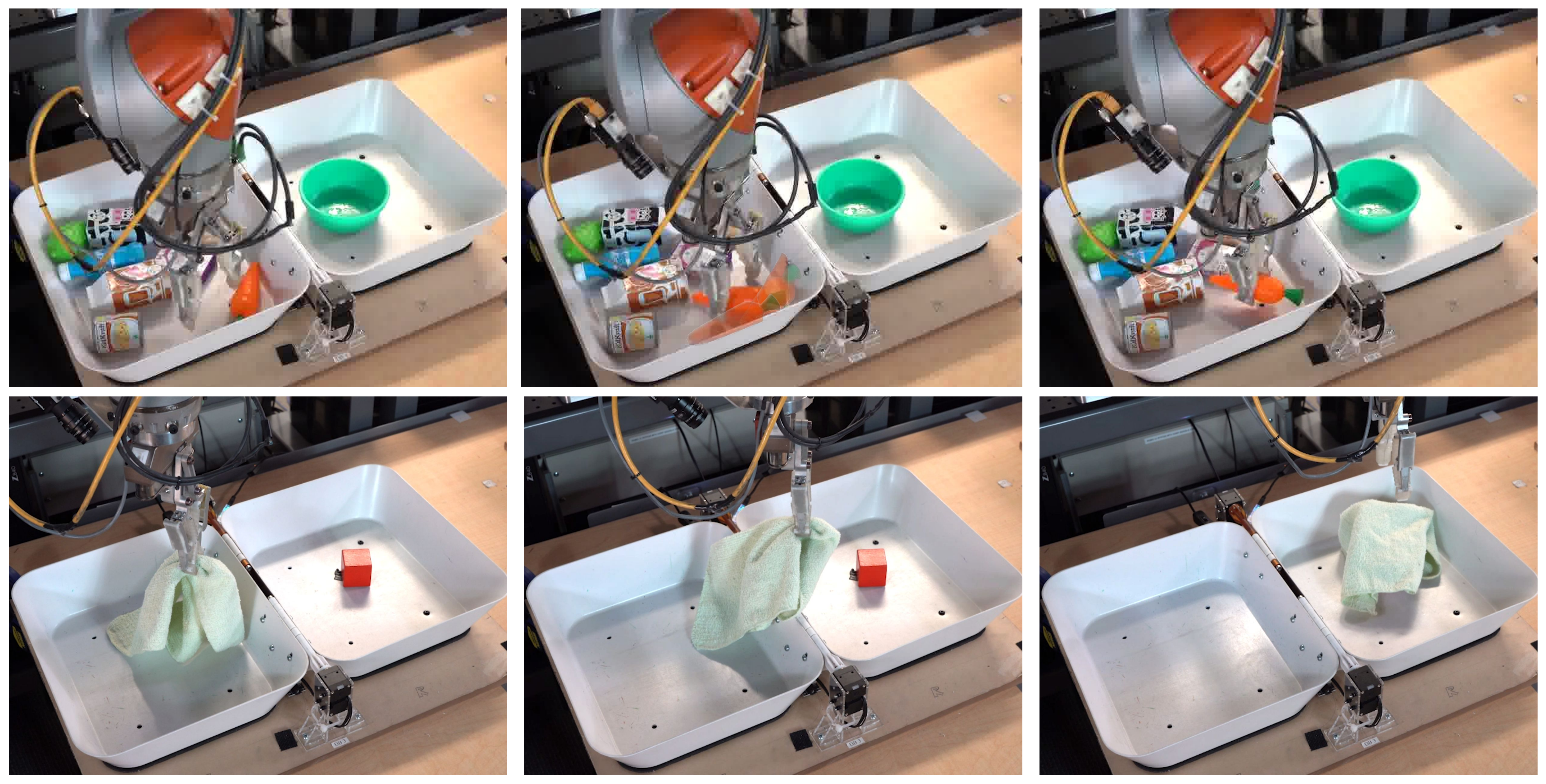}
    \caption{Top row: Example of \textit{pick-carrot}. The robot repositions the carrot out of the corner to pick it. Bottom row: \textit{cover-object}. The deformable cloth is laid over the object.}
    \label{fig:carrot_and_towel}
\end{figure}

MT-Opt can learn broad skills, and then further specialize them to harder but more specific tasks, such as \textit{lift-sausage}.  This retroactive relabelling of prior data is one way to learn new tasks including lifting objects of other properties such as size, location, color or shape.

In addition, MT-Opt can learn new tasks via \textit{proactive} adaptation of known tasks, even ones that are visually and behaviorally different than those in the initial training set.
To demonstrate this, we perform a fine-tuning experiment, bootstrapping from the MT-Opt 12-task model described in Sec.~\ref{sec:mt-opt-quantitative}. 
In particular, we use the MT-Opt policy to collect data for a previously unseen tasks of \textit{lift-cloth} and \textit{cover-object} tasks (see Fig.~\ref{fig:carrot_and_towel} bottom row for an example episode).
Unlike the \textit{lift-sausage} tasks from the above section, prior to starting collection of these new tasks, no episodes in our offline dataset can be relabelled as successes for these two new tasks.

We follow the continuous data collection process described in Sec.~\ref{sec:data_collection}: we define and train the success detector for the new tasks, collect initial data using our \textit{lift-any} and a \textit{place-any} policies, and fine-tune a 14-task MT-Opt model that includes all prior as well as the newly defined tasks. While the new tasks are visually and semantically different, in practice the above mentioned policies give reasonable success rate necessary to start the fine-tuning. We switch to running the new policies on the robots once they are at parity with the \textit{lift-any} and \textit{place-any} policies. After $11K$ \textit{pick-cloth} attempts and $3K$ \textit{cover-object} attempts (requiring $<1$ day of data collection on 7 robots), we obtain an extended 14-task MT policy that performs cloth picking at $70\%$ success and object covering at $44\%$ success. The policy trained only for these two tasks, without support of our offline dataset, yields performance of $33\%$ and $5\%$ respectively, confirming the hypothesis that MT-Opt method is beneficial even if the target tasks are sufficiently different, and the target data is scarce.
By collecting additional $10K$ \textit{pick-cloth} episodes and $6K$ \textit{cover-object} episodes, we further increase the performance of 14-task MT-Opt to $92\%$ and $79\%$, for cloth picking and object covering respectively.
{We perform this fine-tuning procedure with other novel tasks such as previously unseen transparent bottle grasping, which reaches a performance of $60\%$ after less than $4$ days of data collection. } 
Note that in this experiment, we additionally take advantage of the pre-trained MT-Opt policy for collecting the data for the new task. Similarly to other ablations, collecting data using the two-task policy would yield lower success rate per task, leading to larger difference in performance.

%% file: sections/conclusion.tex
\section{Conclusion}
\label{sec:conclusion}

We presented a general multi-task learning framework, MT-Opt, that encompasses a number of elements: a multi-task data collection system that simultaneously collects data for multiple tasks, a scalable success detector framework, and a multi-task deep RL method that is able to effectively utilize the multi-task data.
With real-world experiments, we carefully evaluate various design decisions and show the benefits of sharing weights between the tasks and sharing data using our task impersonation and data re-balancing strategies.
We demonstrate examples of new skills that the system is able to generalize to including placing into new fixtures, covering, aligning, and rearranging.
Finally, we show how MT-Opt can quickly acquire new tasks by leveraging the shared multi-task representations and exploration strategies.

%% file: sections/acknowledgements.tex
\section{Acknowledgements}

The authors would like to thank Josh Weaver, Noah Brown, Khem Holden, Linda Luu and Brandon Kinman for their robot operation support.
We also thank Yao Lu and Anthony Brohan for their help with distributed learning and testing infrastructure;  Tom Small for help with videos and project media; Tuna Toksoz and Garrett Peake for improving the bin reset mechanisms; Julian Ibarz, Kanishka Rao, Vikas Sindhwani and Vincent Vanhoucke for their support; Satoshi Kataoka,  Michael Ahn, and Ken Oslund for help with the underlying control stack; and the rest of the Robotics at Google team for their overall support and encouragement. All of these contributions were incredibly enabling for this project.

%% file: sections/appendix.tex
\clearpage
\section{Appendix}
\label{sec:appendix}

\subsection{Neural Network Architecture}
\label{sec:neural_net}

\begin{figure}[h]
    \centering
    \includegraphics[width=1\linewidth]{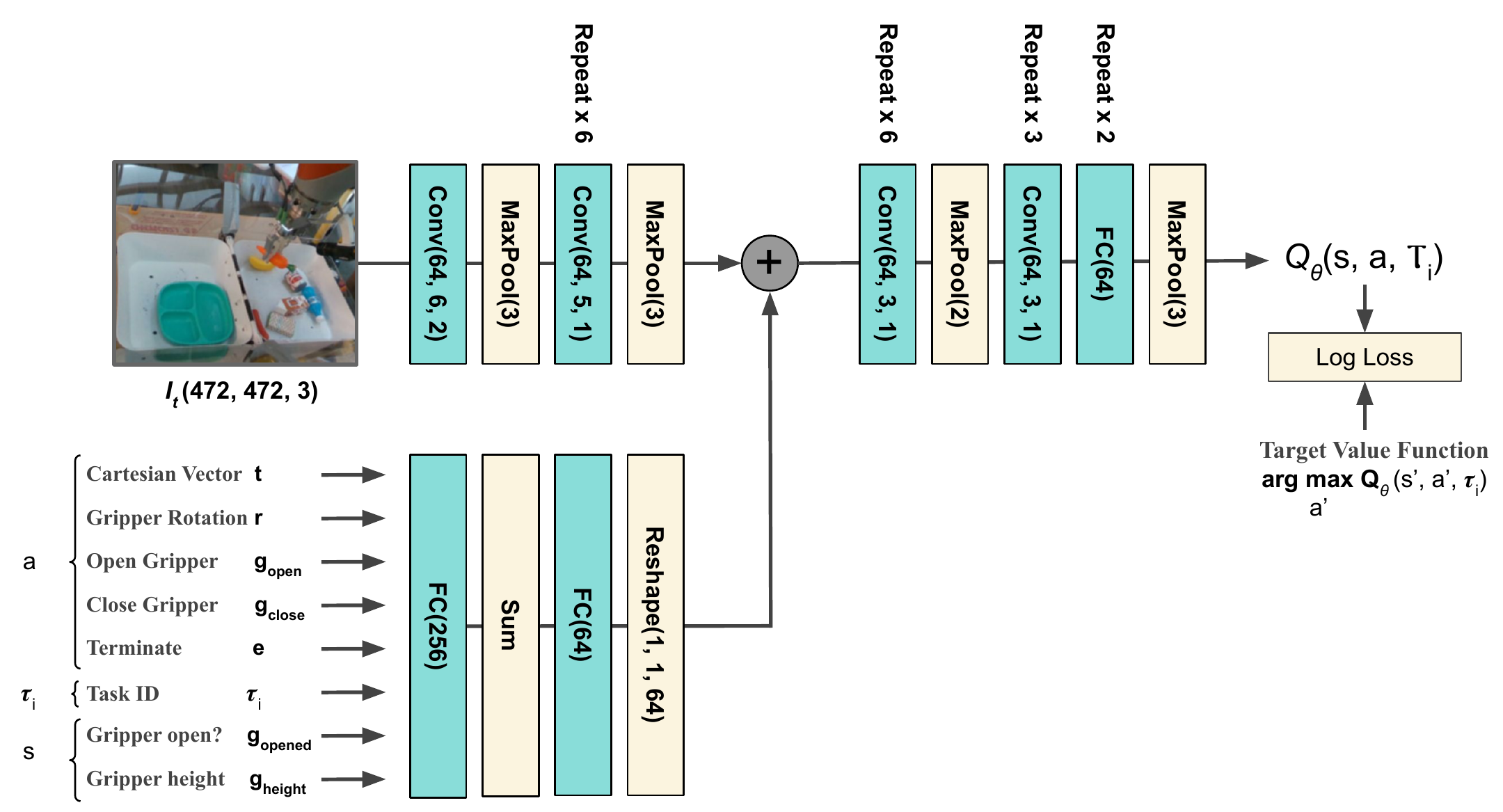}
    \caption{The architecture of MT-Opt Q-function. The input image is processed by a stack of convolutional layers. Action vector, state vector and one-hot vector $\T_i$ representing the task of interest are processed by several fully connected layers, tiled over the width and height dimension of the convolutional map, and added to it. The resulting convolutional map is further processed by a number of convolutional layers and fully connected layers. The output is gated through a sigmoid, such that Q-values are always in the range [0, 1].}
    \label{fig:neural_network}
\end{figure}

We model the Q-function for multiple tasks as a large deep neural network whose architecture is shown in Fig.~\ref{fig:neural_network}. This network resembles one from~\citep{kalashnikov2018qt}. The network takes the monocular RGB image part of the state $s$ as input, and processes it with 7 convolutional layers. The actions $a$ and
additional state features ($g_{status}, g_{height}$) and task ID $\T_i$ are transformed with fully-connected layers, then merged with visual
features by broadcasted element-wise addition. After fusing state and action representations, the Q-value $Q_\theta(s, a)$ is modeled by 9 more convolutional layers followed by two fully-connected layers. In our system the robot can execute multiple tasks from in the given environment. Hence the input image is not sufficient to deduce which task the robot is commanded to execute. To address that, we feed one-hot vector representing task ID into the network to condition Q-Function to learn task-specific control.

\begin{figure}[h]
    \centering
    \includegraphics[width=1\linewidth]{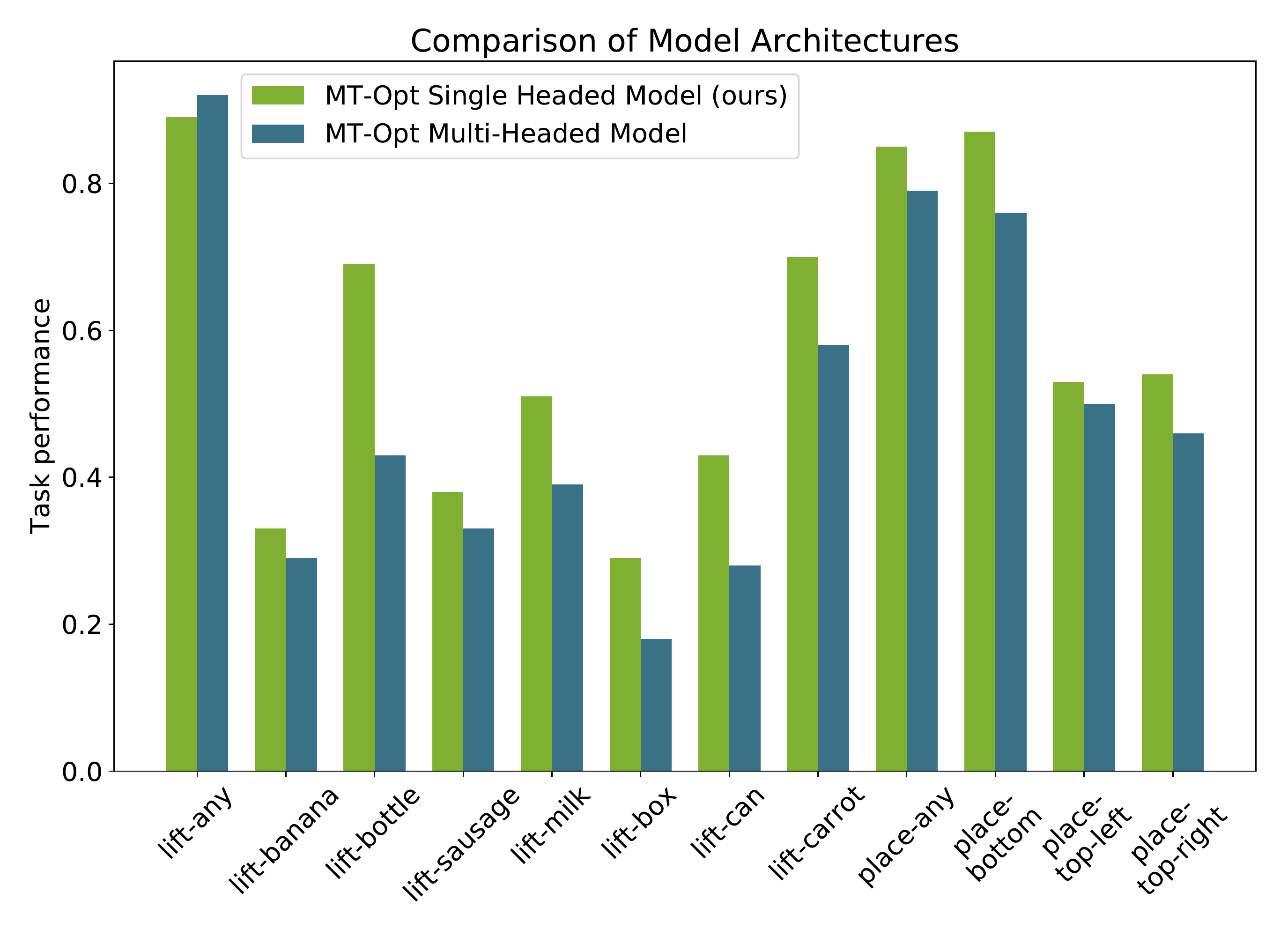}
    \caption{Comparison of single-headed and multi-headed neural networks approximating the Q-function. In both cased task ID was fed as the input to the network. Multi-headed architecture of the Q-function under-performs on a wide range of tasks, winning only on \textit{lift-any} tasks which has most of the data.}
    \label{fig:models_compare}
\end{figure}

In addition to feeding task ID we have experimented with multi-headed architecture, where $n$ separate heads each having 3 fully connected layers representing $n$ tasks were formed at the output of the network. Fig.\ref{fig:models_compare} shows that performance of the system with the multi-headed Q-function architecture is worse almost for all tasks. We hypothesize that dedicated per task heads ``over-compartmentalizes'' task policy, making it harder to leverage shared cross-task representations.

\subsection{Description of Scripted Policies}
\label{subsec:description_of_scripted_policies}
As discussed in Section\ref{sec:data_collection} we use two crude scripted policies to bootstrap easy generic tasks.

\noindent\textbf{Scripted Picking Policy:} To create successful picking episodes, the arm would begin the episode in a random location above the right bin containing objects.  Executing a crude, scripted policy, the arm is programmed to move down to the bottom of the bin, close the gripper, and lift. While the success rate of this policy is very low ($\approx10\%$), especially with the additional random noise injected into actions, this is enough to bootstrap our learning process.\\
\textbf{Scripted Placing Policy:} The scripted policy programmed to perform placing would move the arm to a random location above the left bin that contains a fixture. The arm is then programmed to descend, open the gripper to release the object and retract. This crude policy yields a success rate of (47\%) at the task of placing on a fixture (plate), as the initial fixture is rather large. Data collected by such a simplistic policy is sufficient to bootstrap learning.

\subsection{$f_{I_{skill}}$ impersonation strategy details}
\begin{figure}[t]
    \centering
    \includegraphics[width=0.97\linewidth]{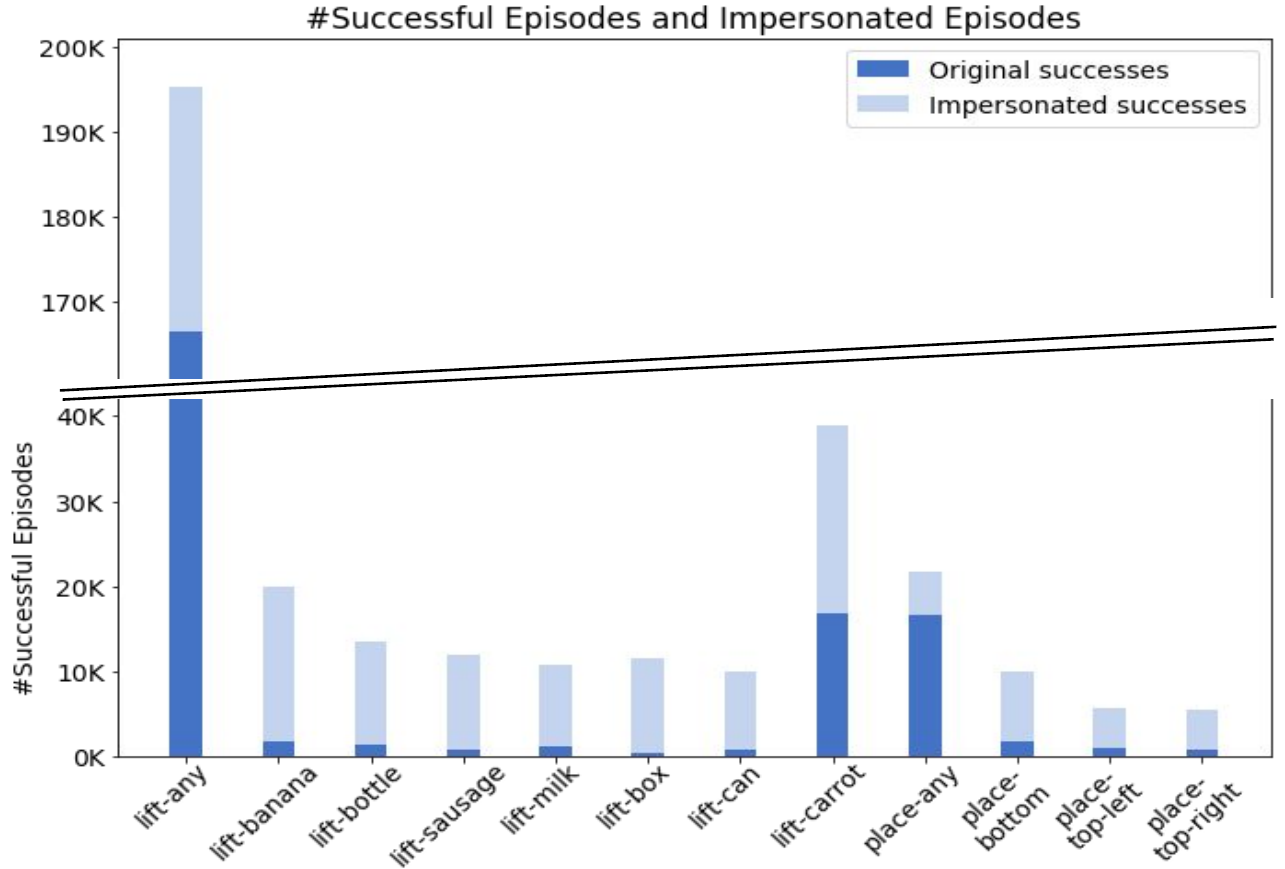}
    
    \vspace{-1mm}
    \caption{Practical effect of task impersonation for successful outcomes. Dark blue indicates data specifically collected for a task; light blue indicates episodes impersonated from some other tasks which happen to be a success for the target task.}
    \label{fig:success_impersonation}
\end{figure}

Task impersonation is an important component of the MT-Opt method. Given an episode and a task definition, the $SD$ classifies if that episode is an example of a successful task execution according to that particular goal definition. Importantly, both the success and the failure examples are efficiently utilized by our algorithm. The success example determines what the task is, while the failure example determines what the task is \textit{not} (thus still implicitly providing the boundary of the task), even if it's an example of a success for some other task. Fig.\ref{fig:success_impersonation} shows by how much the per task data is expanded using the $f_{I_{\text{skill}}}$ impersonation function.

In Section~\ref{sec:tasks-impersonation} we discuss a problem arising when using a naive $f_{I_{all}}$ episodes impersonation function, and suggest a solution to impersonate data only within the boundaries of a skill.
Namely, given an episode $\e_i$ generated by task $\T_i$, a skill $S_j$ that task belongs to is detected. The $e_i$ will be impersonated only for the tasks ${\T_{\{S_j\}}}$ belonging to that particular skill.

\begin{figure*}[h]
    \centering
    \includegraphics[width=0.99\linewidth]{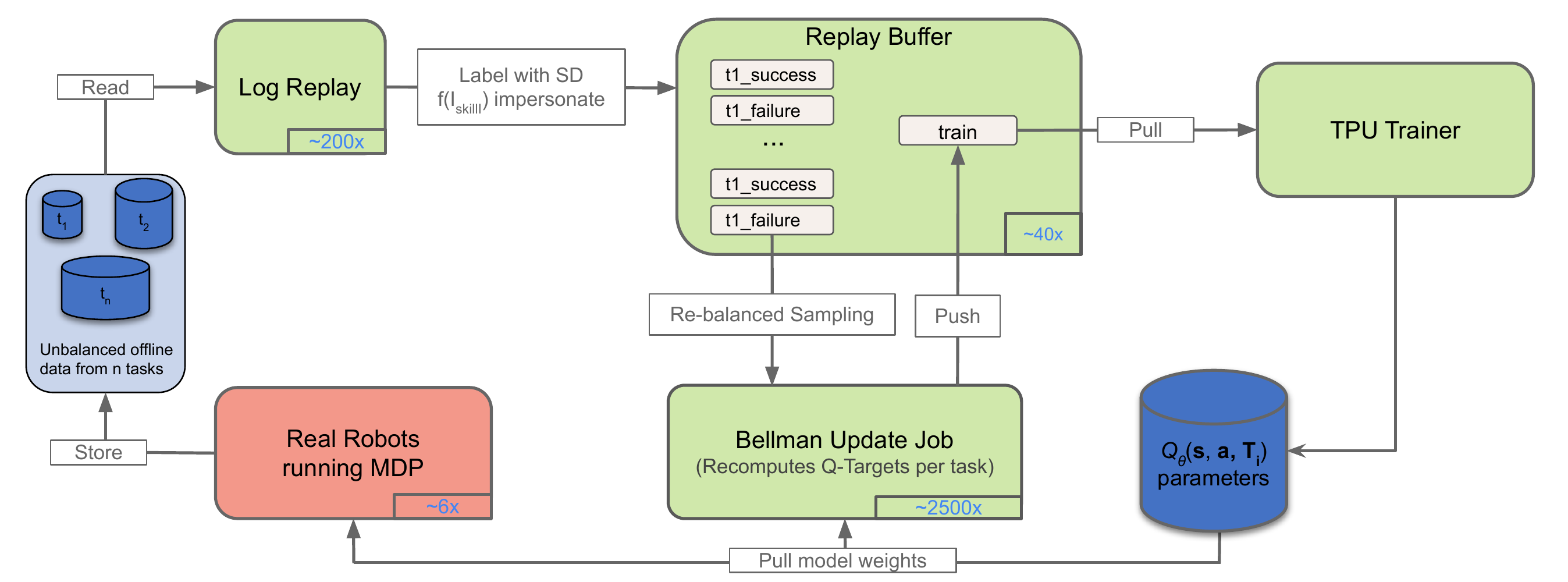}
    \caption{
    \textbf{System overview:} Task episodes from disk are continuously loaded by LogReplay job into task replay buffers. LogReplay process assigns binary reward signal to episodes using available Success Detectors and impersonates episodes using $f_{{I_{skill}}}$ (or other strategy). Impersonated episodes are compartmentalized into dedicated per task buffers, further split into successful and failure groups.  Bellman Update process samples tasks using re-balancing strategy to ensure per task training data balancing and computes Q-targets for individual transitions, which are placed into train buffer.  These transitions ${(\s, \act, \T_i)}$ are sampled by the train workers to update the model weights. The robot fleet and Bellman Update jobs are reloading the most up to date model weights frequently.}
    \label{fig:system_overview}
\end{figure*}

Note, that sometimes impersonation for all ${\T_{\{S_j\}}}$ tasks within a skill could result in too excessive data sharing. For example, the bulk of the data for our \textit{object-acquisition} skill represents variants of tasks involving foods objects. If we want to learn a new task within the same skill using visually significantly different objects, e.g. transparent bottles, all offline episodes involving the plastic objects will be (correctly) impersonated as failures for the \textit{lift-transparent-bottle} task. That is, a few intrinsic failures for that task will be diluted in large set of artificially created negatives.

To solve this issue we introduce a stochastic impersonation function. An impersonated episode candidate will be routed to training with the probability $p_s$ if it's a success, or with probability $p_f$ if it's a failure. We experiment with $p_s=1.0$, and $p_f<=1.0$. The reasoning is that it’s always desirable to utilize surplus impersonated examples of a \textit{successful} task execution, but it could be better to utilize only a fraction of the surplus \textit{failures} to balance intrinsic v.s. artificial failures for that task.

This gives rise to the $f_{I_{skill}}(p_s, p_f)$ impersonation function which is suitable in some situations explained above.

\subsection{Distributed Asynchronous System}
\label{sec:distributed}
Fig.\ref{fig:system_overview} provides an overview of our large scale distributed Multi-Task Reinforcement Learning system.

\section{Reward Specification with Multi-Task Success Detector}
\label{sec:sd-details}
\begin{table*}[]
\centering
\begin{tabular}{|c|c|c|c|c|c|c|c|c|}
\hline
\begin{tabular}[c]{@{}c@{}}Primary Success \\ Detector Name\end{tabular} & \begin{tabular}[c]{@{}c@{}}Total \\ Count\end{tabular} & \begin{tabular}[c]{@{}c@{}}Success\\  Count\end{tabular} & \begin{tabular}[c]{@{}c@{}}Failure \\ Count\end{tabular} & \begin{tabular}[c]{@{}c@{}}Success \\ Rate\end{tabular} & \begin{tabular}[c]{@{}c@{}}False Negative \\ Rate\end{tabular} & \begin{tabular}[c]{@{}c@{}}False Positive\\  Rate\end{tabular} & \begin{tabular}[c]{@{}c@{}}Other Task \\ False Negative Rate\end{tabular} & \begin{tabular}[c]{@{}c@{}}Other Task \\ False Positive Rate\end{tabular} \\ \hline
\textit{lift-any}                                                        & 16064                                                  & 7395                                                   & 8672                                                   & 46\%                                                    & 1\%                                                            & 2\%                                                            & 0\%                                                                       & 0\%                                                                       \\ \hline
\textit{lift-banana}                                                     & 6255                                                   & 510                                                    & 5745                                                   & 8\%                                                     & 2\%                                                            & 1\%                                                            & 0\%                                                                       & 1\%                                                                       \\ \hline
\textit{lift-bottle}                                                     & 6472                                                   & 430                                                    & 6042                                                   & 7\%                                                     & 5\%                                                            & 1\%                                                            & 0\%                                                                       & 1\%                                                                       \\ \hline
\textit{lift-sausage}                                                    & 6472                                                   & 461                                                    & 6011                                                   & 7\%                                                     & 3\%                                                            & 0\%                                                            & 0\%                                                                       & 1\%                                                                       \\ \hline
\textit{lift-milk}                                                       & 6472                                                   & 158                                                    & 6314                                                   & 2\%                                                     & 7\%                                                            & 0\%                                                            & 3\%                                                                       & 9\%                                                                       \\ \hline
\textit{lift-box}                                                        & 6467                                                   & 487                                                    & 5980                                                   & 8\%                                                     & 1\%                                                            & 1\%                                                            & 0\%                                                                       & 2\%                                                                       \\ \hline
\textit{lift-can}                                                        & 6467                                                   & 270                                                    & 6197                                                   & 4\%                                                     & 2\%                                                            & 0\%                                                            & 3\%                                                                       & 3\%                                                                       \\ \hline
\textit{lift-carrot}                                                     & 6481                                                   & 911                                                    & 5570                                                   & 14\%                                                    & 0\%                                                            & 1\%                                                            & 0\%                                                                       & 0\%                                                                       \\ \hline
\textit{place-any}                                                       & 3087                                                   & 1363                                                   & 1724                                                   & 44\%                                                    & 1\%                                                            & 2\%                                                            & 0\%                                                                       & 0\%                                                                       \\ \hline
\textit{place-bottom}                                                    & 2893                                                   & 693                                                    & 2200                                                   & 24\%                                                    & 2\%                                                            & 1\%                                                            & 1\%                                                                       & 3\%                                                                       \\ \hline
\textit{place-top-left}                                                  & 2895                                                   & 346                                                    & 2549                                                   & 12\%                                                    & 10\%                                                           & 0\%                                                            & 3\%                                                                       & 8\%                                                                       \\ \hline
\textit{place-top-right}                                                 & 2897                                                   & 312                                                    & 2585                                                   & 11\%                                                    & 4\%                                                            & 0\%                                                            & 0\%                                                                       & 5\%                                                                       \\ \hline
\end{tabular}
\caption{Success detection holdout data statistics. Table shows success detector error rate for held out labelled success detector data. We split out the evaluation dataset based on the robot, e.g. all data generated by Robot \#1 is used for evaluations and not for training. This strategy results in a much better test of generalization power of the success detector, compared to the conventional way to split out $20\%$ of the data randomly for evaluation.
The Other Task False [Positive/Negative] Rates columns indicates how well the success detector for a task A classifies outcomes for all other tasks. For example we want to ensure that a successful \textit{lift-carrot} episode does not trigger \textit{lift-banana} success, i.e. not only a success detector should manifest its dedicated task success, but also reliably reason about other related tasks. This ``contrastiveness'' property of the success detectors is of great importance in our system. As success detectors determine tasks data routing and experience sharing, an error in this tasks data assignment would drive anti-correlated examples for each task, resulting in a poor performance of the system.}
\label{Table:sd_stats}
\end{table*}

\begin{figure}[h]
    \centering
    \includegraphics[width=1\linewidth]{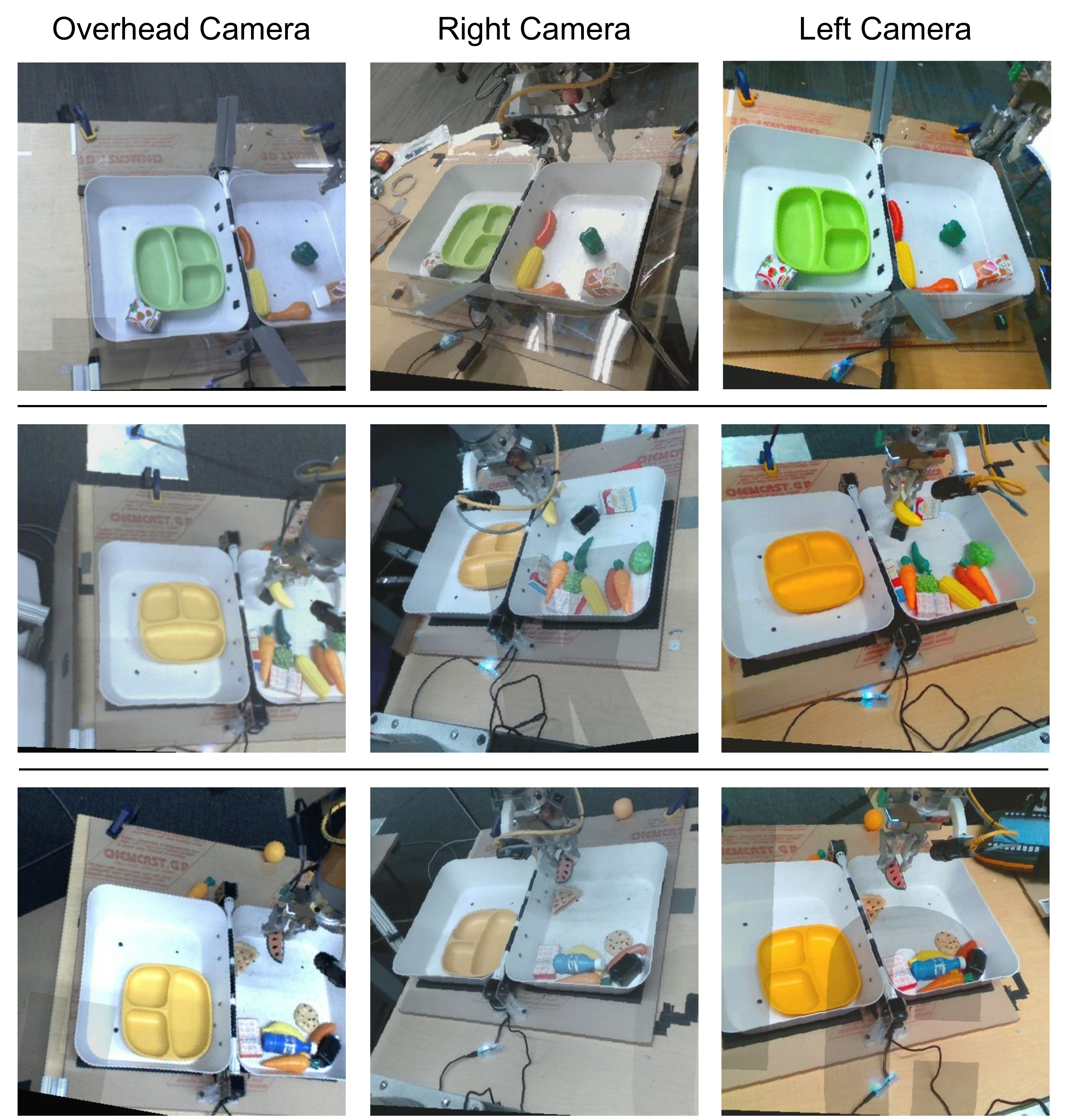}
    \caption{$SD$ training images. Each row represents a set of images captured at the same time that are fed into the $SD$ model.  These images demonstrate our train-time $SD$ data augmentation process as they have been distorted via cropping, brightening, rotating, and superimposing of shadows.}
    \label{fig:sd_images}
\end{figure}
Training a visual success detector is an iterative process, as a new task initially has no data to train from. We have two strategies to efficiently create an initial $SD$ training dataset for a new task. 1) We collect 5Hz videos from 3 different camera angles where every frame of the video a human is demonstrating task success, and then a short video demonstrating failure. Note that the user shows the desired and non-desired \textit{outcome} of the task, not to be confused with demonstrations of how the task needs to be done. The user would then change the lighting, switch out the objects and background, and then collect another pair of example videos (see Fig.~\ref{fig:sd_video_images} for example one video where there is always something on a plate being moved around paired with another video where there is never anything on a plate). The intention here is to de-correlate spurious parts of the scene from task-specifics. This process is repeated for approximately 30 minutes per task. 2) We relabel data from a policy that occasionally generated success for the new task (e.g., relabel \textit{lift-any} data for \textit{lift-carrot} task.). 

Once the initial $SD$ is trained, we can train an RL policy, and begin on-policy collection.  We continue to label on-policy data which keeps coming for the new task until $SD$ is reliable.  Table~\ref{Table:sd_stats} shows false positive and false negative error rates on holdout data for the $SD$ model used in our ablations. Our holdout data consisted of all images from a particular robot.

During the $SD$ training process, the data is artificially augmented to improve generalization, which involves cropping, brightening, rotating, and superimposing random shadows onto the images. Fig.~\ref{fig:sd_images} shows training images after these distortions have been applied.
Our success detector model is trained using supervised learning, where we balance the data between success and failures as well as tasks. We use the architecture that is based on that from~\cite{akinola2020learning} with the exception of the action conditioning as it is not needed for this classification task. For each task the network outputs the probability representing whether a given state was a success or failure for the corresponding task.  The model receives three images as an input that come from an over-the-shoulder camera (same image as RL policy), and two additional side cameras.  These side camera images are only used by the $SD$ model, not the RL model.  The additional cameras ensured that the task goals would be unambiguous, with a single camera, it was often difficult for a human to discern from an image whether or not the task had succeeded.

\begin{figure}[h]
    \centering
    \includegraphics[width=1\linewidth]{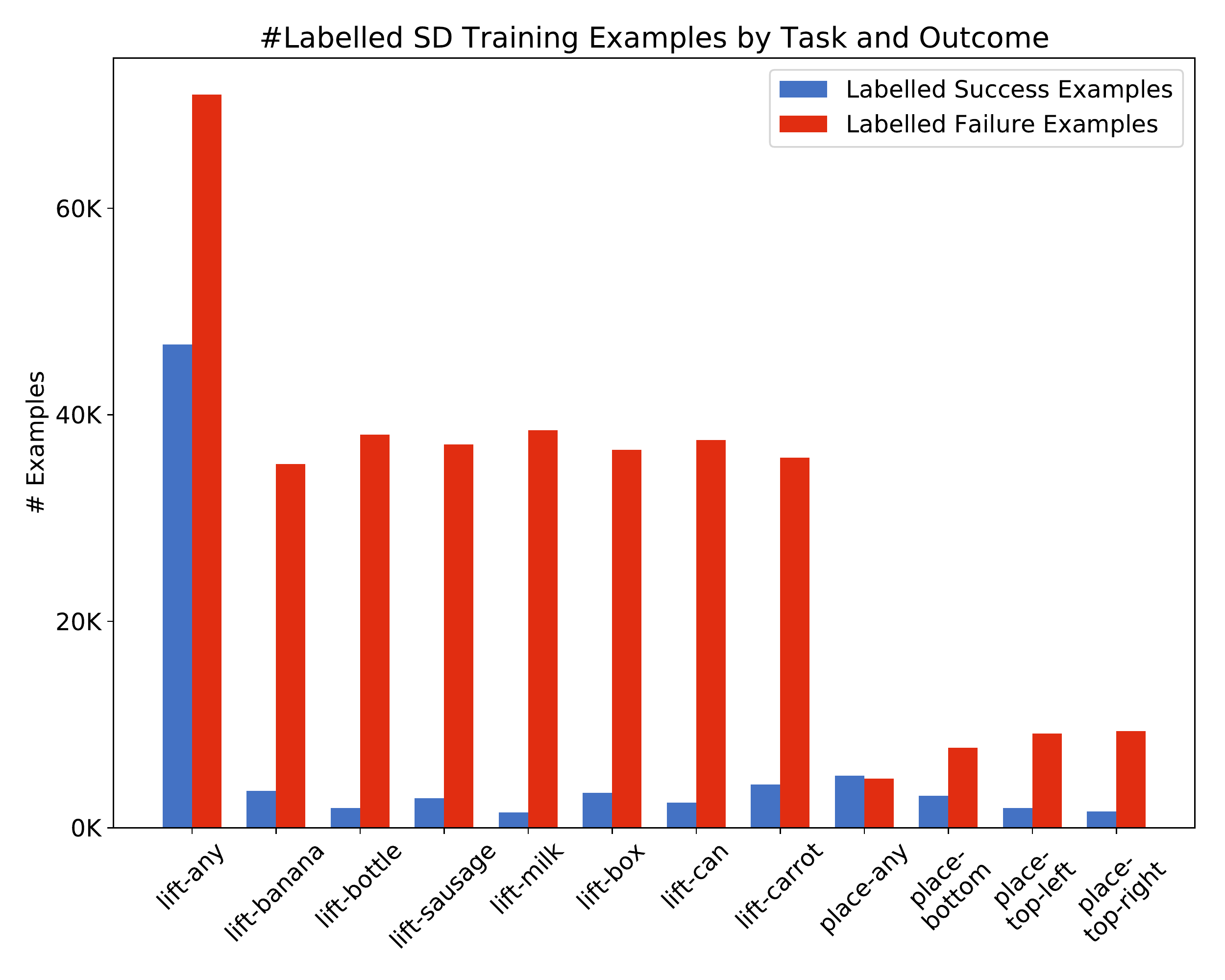}
    \vspace{-7mm}
    \caption{Counts of labelled $SD$ training data by task and outcome.  This data was generated either from human video demonstration, or by labelling terminal images from episodes produced by a robot. Note, that not all of the negatives were hand-labelled. As we may know dependencies between the tasks, e.g. that a success for \textit{lift-carrot} is always a failure for \textit{lift-banana}, we can automatically generate negative examples. Similarly, all successes for the semantic lifting tasks are also successes for the \textit{lift-any} task.}
    \label{fig:sd_train_data_stats}
\end{figure}

A breakdown of the labelled $SD$ training data is provided in Fig.~\ref{fig:sd_train_data_stats}. 
While training SD, we incorporated data sharing logic based on task feasibility. For example any success for \textit{lift-carrot} would also be marked as failure for all other instance lifting tasks, and as a success for \textit{lift-any}.  In this manner, the original set of labelled data shown in Fig. \ref{fig:sd_train_data_stats} could act effectively as a much larger dataset for all tasks, where successes of one task often worked an interesting negatives for other tasks.  Additionally we balanced the proportion of success and failure examples per task seen by the model during training.

\input{sections/robotic_setup}

\input{sections/data_collection_details}

\input{sections/full_real_ablations}

%% file: sections/robotic_setup.tex
\section{Robot Setup}
\label{sec:real_world_setup}

In order for our system to be able to learn a vision-based RL policy that can accomplish multiple tasks, we need to collect a large, diverse, real-robot dataset that represents data for various tasks. 

To achieve this goal, we set up an automated, multi-robot data collection system where each robot picks a task $\T_i$ to collect the data for. Collected episode is stored on disk along with the $\T_i$ bit of information. Our learning system can then use this episode collected $\T_i$ for to train a set of other tasks utilizing MT-Opt data impersonation algorithm. Once the episode is finished, our data collection system decides whether to continue with another task or perform an automated reset of the workspace.

\begin{figure}[h]
    \centering
    \includegraphics[width=1\linewidth]{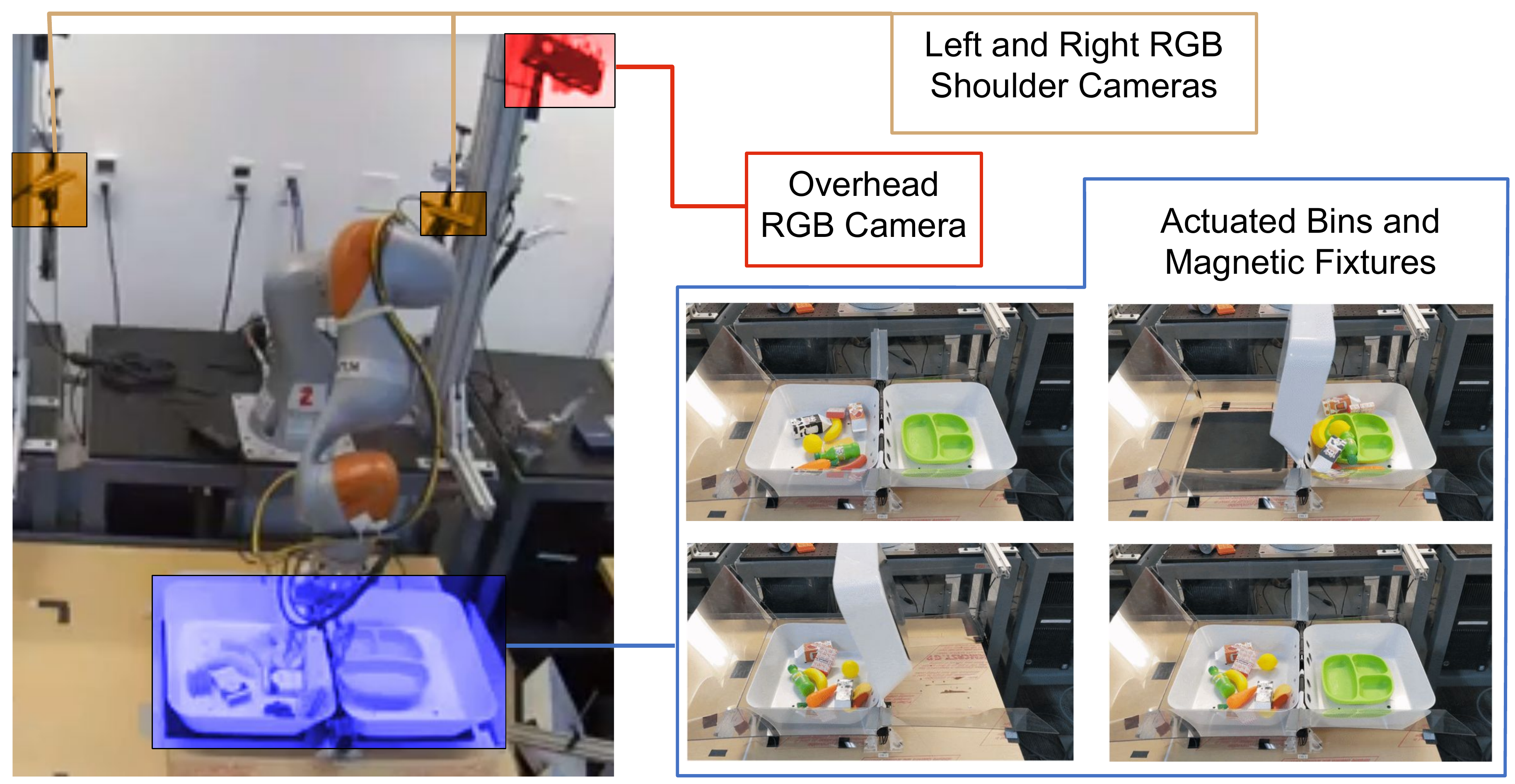}
    \caption{Robot workspace consisting of an overhead camera (red), two over the shoulder cameras (brown), and a pair of articulated resettable bins with a plate fixture that can be magnetically attached to the bin (blue).}
    \label{fig:robot_workspace}
\end{figure}

In particular, we utilize 7 KUKA IIWA arms with two-finger grippers and 3 RGB cameras (left, right, and over the shoulder). In order to be able to automatically reset the environment, we create an actuated resettable bin, which further allows us to automate the data collection process. More precisely, the environment consists of two bins (with the right bin containing all the source objects and the left bin containing a plate fixture magnetically attached anywhere on the workbench) that are connected via a motorized hinge so that after an episode ends, the contents of the workbench can be automatically shuffled and then dumped back into the right bin to start the next episode. Fig.~\ref{fig:robot_workspace} depicts the physical setup for data collection and evaluation. This data collection process allows us to collect diverse data at scale: 24 hours per day, 7 days a week across multiple robots.

One episode has $\approx10$ steps on average, taking $\approx25$ seconds to be generated on a robot, including environment reset time. This accounts to $\approx3300$ episodes/day collected on a single robot, or $\approx23K$ episodes/day collected across our fleet of 7 robots.

\begin{figure}[h]
    \centering
    \includegraphics[width=0.7\linewidth]{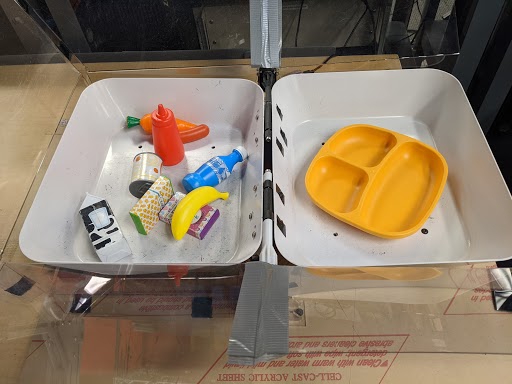}
    \caption{Evaluation scene used for ablation experiments. Contains one of three different color plates.  And nine graspable objects: One of each object from our seven object categories with two extra toy food objects sometimes from the seven object categories, sometimes not.}
    \label{fig:eval_scene}
\end{figure}

%% file: sections/data_collection_details.tex
\subsection{Details of Data Collection to bootstrap a Multi-Task System}
\label{sec:data_collection_details}

\begin{table*}
\centering
\begin{tabular*}{\textwidth}{l|c|c|c|c|c|c|c|c|c}

Task Name  & \#Eps. & QT-Opt & $f_{I_{\text{orig}}}, \text{rand}$ & $f_{I_{\text{orig}}}, \text{rebal}$ & $f_{I_{\text{all}}}, \text{rand}$ & $f_{I_{\text{all}}}, \text{rebal}$ &  $f_{I_{\text{skill(1, 0.15)}}}, \text{rebal}$ &
$f_{I_{\text{skill(1, 1)}}}, \text{rand}$ &
$f_{I_{\text{skill(1, 1)}}}, \text{rebal}$\\
&  & & $\text{QT-Opt}$ & & $\text{DataShare}$ & & & & $\text{ (ours)}$\\
&  & & $\text{MultiTask}$ & & $\text{MultiTask}$ & & & & $\text{}$\\
\hline
\textit{lift-any} & 635K & 0.88 & 0.94 & 0.85 & 0.62 & $\textbf{0.95}$ & 0.80 & 0.88 & 0.89\\
\textit{lift-banana} & 9K & 0.04 & 0.13 & 0.38 & 0.09 & 0.30 & 0.58 & $\textbf{0.62}$ & 0.33\\
\textit{lift-bottle} & 11K & 0.02 & 0.16 & 0.66 & 0.15 & 0.48 & 0.68 & 0.55 & $\textbf{0.69}$\\
\textit{lift-sausage} & 5K & 0.02 & 0.10 & 0.38 & 0.15 & 0.39  & $\textbf{0.42}$ & 0.28 & 0.38\\
\textit{lift-milk} & 6K & 0.01 & 0.13 & 0.42 & 0.13 & 0.27 & 0.27 & $\textbf{0.52}$ & 0.51\\
\textit{lift-box} & 6K & 0.00 & 0.12 & 0.16 & 0.08 & 0.22 & 0.12 & 0.28 & $\textbf{0.29}$\\
\textit{lift-can} & 6K & 0.01 & 0.16 & 0.46 & 0.07 & 0.28 & $\textbf{0.47}$ & 0.43 & 0.43\\
\textit{lift-carrot} & 80K & 0.71 & 0.41 & 0.72 & 0.37 & $\textbf{0.75}$ & 0.52 & 0.71 & 0.70\\
\textit{place-any} & 30K & N/A & $\textbf{0.86}$ & 0.74 & 0.30 & 0.24 & 0.83 & 0.57 & 0.85\\
\textit{place-bottom} & 5K & N/A & 0.43 & 0.57 & 0.30 & 0.02 & 0.62 & 0.17 & $\textbf{0.87}$\\
\textit{place-top-right} & 4K & N/A & 0.16 & $\textbf{0.55}$ & 0.08 & 0.10 & 0.26 & 0.27 & 0.54\\
\textit{place-top-left} & 4K & N/A & 0.23 & $\textbf{0.75}$ & 0.19 & 0.16 & 0.39 & 0.22 & 0.53\\
\hline
Min & & 0.00 & 0.10 & 0.16 & 0.07 & 0.02 & 0.12 & 0.17 & $\textbf{0.29}$\\
25-th percentile & & 0.00 & 0.13 & 0.41 & 0.09 & 0.20 & 0.36 & 0.28 & $\textbf{0.42}$\\
Median & & 0.01 & 0.16 & $\textbf{0.56}$ & 0.15 & 0.28 & 0.50 & 0.48 & 0.54\\
Mean & & 0.14 & 0.32 & 0.55 & 0.21 & 0.35 & 0.49 & 0.46 & $\textbf{0.58}$\\
75-th percentile & & 0.03 & 0.42 & 0.73 & 0.30 & 0.41 & 0.64 & 0.58 & $\textbf{0.74}$\\
Max & & 0.88 & 0.94 & 0.85 & 0.62 & $\textbf{0.95}$ & 0.83 & 0.88 & 0.89\\
Mean (low data) & & 0.01 & 0.18 & 0.42 & 0.13 & 0.21 & 0.36 & 0.32 & $\textbf{0.5}$\\

\end{tabular*}
\caption{Quantitative evaluation of MT-Opt with different data impersonation and re-balancing strategies. This table reports performance of 7 different models on the 12 ablation tasks, trained on identical offline dataset, with identical computation budget, and evaluated executing 100 attempts for each task for each strategy on the real robots (totaling to 12*100*7=8400 evaluations). In all cases a shared policy for all 12 tasks is learned. The difference across the strategies is in the way the data is impersonated (expanded), and in the way the impersonated data is further re-balanced.  The last column is our best strategy featuring skill-level data impersonation and further data re-balancing. This strategy outperforms other strategies on many different percentiles across all 12 tasks; however the effect of that strategy is even more pronounced for the tasks having scarce data, e.g. \textit{lift-can}, \textit{lift-box}, \textit{place-top-right}, see Mean (low data) statistic. The column \#2 indicates the number of episodes which were collected for each task.}
\label{tab:full_ablation_results}
\end{table*}

This section contains more details on the data collection process introduced in Section~\ref{sec:data_collection_details}. \underline{Real world robot data is noisy}. For this project nearly 800,000 episodes were collected through the course of 16 months. The data was collected over different:
\begin{enumerate}
    \item Locations: Three different physical lab locations.
    \item Time of day: Robots ran as close to 24x7 as we could enable.
    \item Robots: 6-7 KUKAs with variations in background, lighting, and slight variation in camera pose.
    \item Success Detectors: We iteratively improved our success detectors.
    \item RL training regimes: We developed better training loops hyper-parameters and architectures as time went on.
    \item Policies: Varied distribution of scripted, epsilon greedy, and on-policy data collection over time.
\end{enumerate}

\begin{figure}[h]
    \centering
    \includegraphics[width=0.99\linewidth]{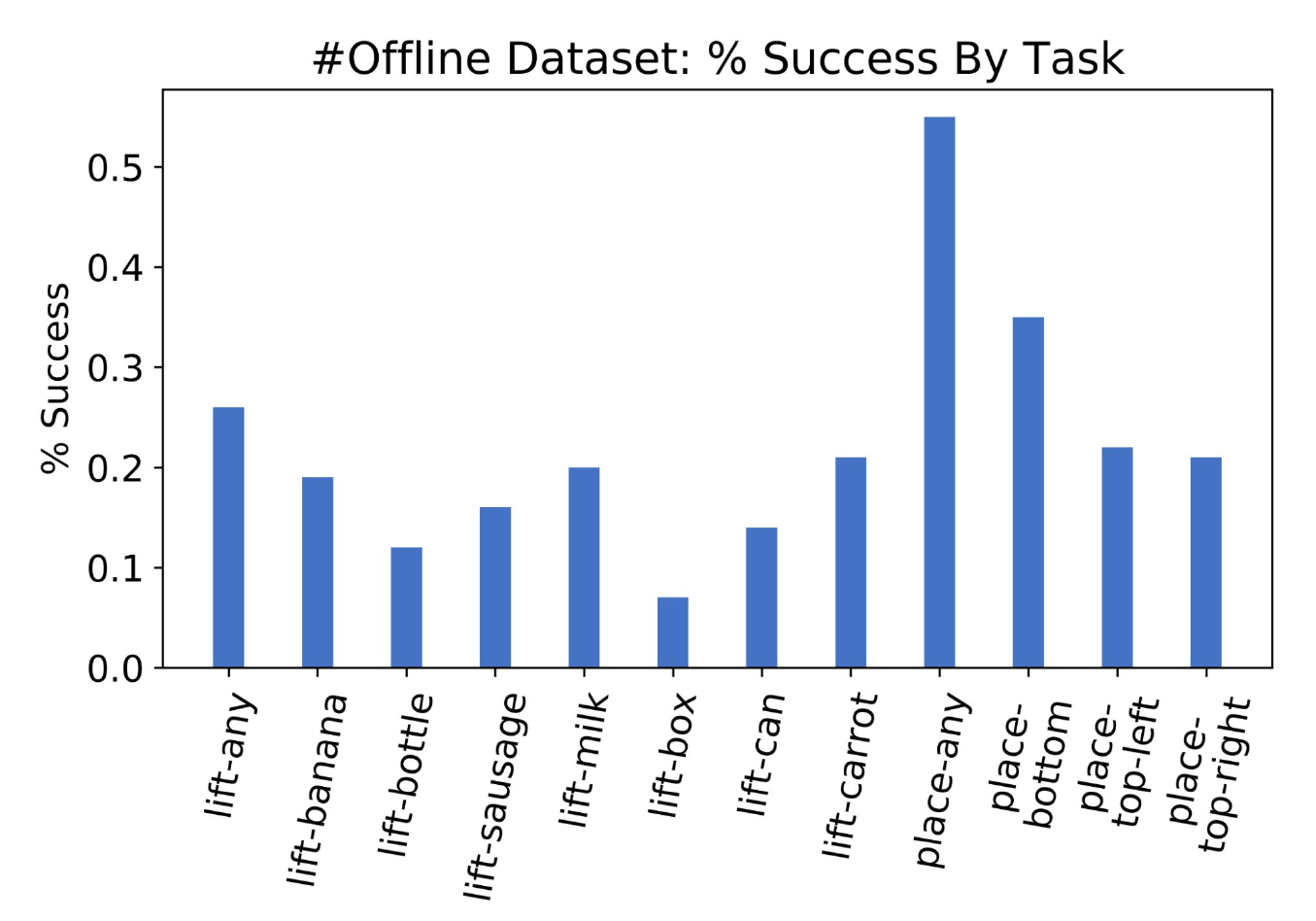}
    \caption{Effective success rate for each task in our offline dataset.  This plot represents the distribution of successes within the entirety of our offline dataset collected over time from many policies, not the performance of any particular policy.}
    \label{fig:offline_dataset_success_rate}
    \vspace{-8pt}
\end{figure}

Our data collection started in an original physical lab location, was paused due to COVID-19, and the robots were later setup at a different physical lab location affecting lighting and backgrounds. Initially scripted policies were run collecting data for the \textit{lift-anything} and \textit{place-anywhere} tasks. Once performance of our learned policy for these tasks out-performed the scripted policy we shifted to a mix of epsilon greedy and pure on-policy data collection. The majority of our episodes were collected for the \textit{lift-anything} and \textit{place-anywhere} tasks with learned policies. It is worth mentioning that over the course of data collection many good and bad ideas where tried and evaluated via on-policy collection. All of these episodes are included in our dataset. Additional tasks being incorporated over time.   

After we had a policy capable of the \textit{lift-anything} and \textit{place-anywhere} tasks we introduced more specific variations of pick and place tasks where either a specific object needed to be picked, or an object needed to be placed in a specific location on the plate. At this point, our data collection process consisted of executing a randomly selected pick task followed by a randomly selected place task. 

As a result of the collection process described above, we were left with a 800,000+ episode offline dataset, very diverse along tasks, policies, success rate dimensions.

%% file: sections/full_real_ablations.tex
\section{Details for real world experiments}

The robot workspace setup for the 12 task ablations is shown in Fig ~\ref{fig:eval_scene}. Table~\ref{tab:full_ablation_results} summarizes studies of 7 different data impersonation and re-balancing strategies for 12 tasks. The last column features the model which on average outperforms other strategies. Note that this strategy is not the best across the board. For example, due to big imbalance of our offline dataset, the native data management strategy (column \#3) yields best performance for the over represented tasks, but very bad performance for underrepresented tasks.